\documentclass[10pt,twocolumn,letterpaper]{article}
\usepackage[T1]{fontenc}
\usepackage[pagenumbers]{cvpr}
\input{arxiv_layout}
\usepackage{amsmath,amssymb,bm}
\usepackage{graphicx,booktabs,multirow,array,xspace}
\usepackage{colortbl}
\definecolor{cvprblue}{RGB}{40,97,138}
\definecolor{draftgray}{RGB}{85,97,110}
\definecolor{tabblue}{RGB}{235,242,247}
\definecolor{tablebest}{HTML}{FF0000}
\definecolor{tablesecond}{HTML}{00B0F0}
\definecolor{tablehead}{HTML}{E6E6E6}
\definecolor{tableours}{HTML}{FFF8F1}
\newcommand{\bestscore}[1]{\textcolor{tablebest}{\textbf{#1}}}
\newcommand{\secondscore}[1]{\textcolor{tablesecond}{\textbf{#1}}}
\usepackage[pagebackref,breaklinks,colorlinks,allcolors=cvprblue]{hyperref}
\DeclareMathOperator*{\argmax}{arg\,max}
\newcommand{\method}{{\upshape\textsc{CamoBreak}}\xspace}
\newcommand{\CSP}{\operatorname{CSP}}

\newcommand{\compacttable}{\fontsize{8}{9}\selectfont
  \setlength{\tabcolsep}{3pt}\renewcommand{\arraystretch}{.96}
  \setlength{\aboverulesep}{1.2pt}\setlength{\belowrulesep}{1.2pt}}
\newsavebox{\tablebox}
\newcommand{\fullwidthtable}[2]{\sbox{\tablebox}{#2}%
  \setlength{\tabcolsep}{\dimexpr\tabcolsep+(\textwidth-\wd\tablebox)/#1\relax}%
  #2}
\newcommand{\fittable}[2]{\sbox{\tablebox}{#2}%
  \ifdim\wd\tablebox>#1\resizebox{#1}{!}{\usebox{\tablebox}}%
  \else\usebox{\tablebox}\fi}
\newlength{\tablepadding}
\newcommand{\columntable}[2]{\begingroup
  \sbox{\tablebox}{#2}%
  \setlength{\tablepadding}{\dimexpr\tabcolsep+(\linewidth-\wd\tablebox)/#1/2\relax}%
  \ifdim\tablepadding<1pt
    \setlength{\tabcolsep}{1pt}\sbox{\tablebox}{#2}%
    \resizebox{\linewidth}{!}{\usebox{\tablebox}}%
  \else
    \setlength{\tabcolsep}{\tablepadding}\sbox{\tablebox}{#2}%
    \setlength{\tabcolsep}{\dimexpr\tabcolsep+(\linewidth-\wd\tablebox)/#1/2\relax}%
    \sbox{\tablebox}{#2}%
    \resizebox{\linewidth}{!}{\usebox{\tablebox}}%
  \fi\endgroup}

\newcommand{\R}{\mathbb{R}}

\usepackage{placeins,needspace}
\usepackage{flushend}
\makeatletter
\newbox\balance@outer
\newbox\balance@inner
\newcommand{\balance@unwrap}[1]{%
  \setbox\balance@outer\vbox{\unvcopy#1\global\setbox\balance@inner\lastbox}%
  \ifvbox\balance@inner
    \ifdim\dimexpr\ht\balance@outer+\dp\balance@outer\relax=0pt
      \setbox#1\vbox{\unvbox\balance@outer\unvbox\balance@inner}%
    \fi
  \fi}
\pretocmd{\last@outputdblcol}{%
  \balance@unwrap\@outputbox
  \if@firstcolumn\else\balance@unwrap\@leftcolumn\fi
}{}{\PackageError{camobreak}{Column balancing patch failed}{Check flushend compatibility.}}
\newcommand{\balance@mark}[1]{%
  \protected@write\@auxout{}{\string\def\string\@LN@column{#1}}}
\patchcmd{\last@outputdblcol}{\fmt@box@adds{\box\@leftcolumn}}%
  {\balance@mark{1}\fmt@box@adds{\box\@leftcolumn}}{}{}
\patchcmd{\last@outputdblcol}{\fmt@box@adds{\box\@outputbox}}%
  {\balance@mark{2}\fmt@box@adds{\box\@outputbox}}{}{}
\makeatother
\title{Look Where Camouflage Breaks:\\Concealment-Guided Observation for Camouflaged Object Detection}
\author{Dingming Zhang$^{1}$,\;
Fengyang Xiao$^{1}$,\;
Jingjia Feng$^{1}$,\;
Chunming He$^{1,\dagger}$,\;
Sina Farsiu$^{1,\dagger}$\\
{\normalsize $^{1}$Duke University}\\
{\footnotesize $\dagger$ Corresponding authors: chunming.he@duke.edu / sina.farsiu@duke.edu}}
\hypersetup{
  pdftitle={Look Where Camouflage Breaks: Concealment-Guided Observation for Camouflaged Object Detection},
  pdfauthor={Dingming Zhang, Fengyang Xiao, Jingjia Feng, Chunming He, Sina Farsiu}
}
\makeatletter
\patchcmd{\@maketitle}{\vspace*{24pt}}{\vspace*{4pt}}{}{
  \PackageError{camobreak}{Title spacing patch failed}{Check the title definition.}}
\patchcmd{\@maketitle}{\vspace*{12pt}}{\vspace*{4pt}}{}{
  \PackageError{camobreak}{Author spacing patch failed}{Check the title definition.}}
\g@addto@macro\@maketitle{
\begin{minipage}{\textwidth}
\centering
\includegraphics[width=\textwidth]{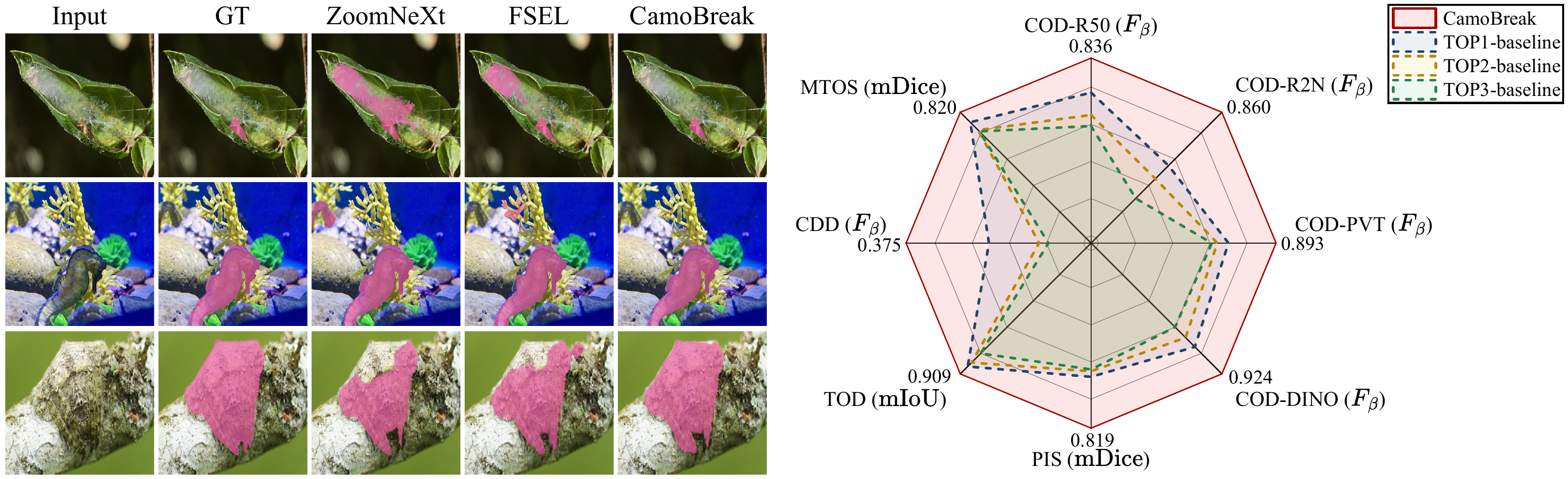}
\captionsetup{skip=2pt}
\captionof{figure}{\textbf{Concealment-guided observation.} \method improves target recovery across encoder families and tasks. COD scores average adaptive $F_\beta$ over all test images across the four benchmarks. Each TOP$k$ curve connects the $k$-th best baseline on each task axis.}
\label{fig:teaser}
\end{minipage}
\par\vspace{5pt}
}
\makeatother
\begin{document}
\AddToShipoutPicture{%
  \AtTextUpperLeft{%
    \put(0,\LenToUnit{36pt}){%
      \normalfont\rmfamily\small\color{black}Preprint. Under review.}%
    \put(0,\LenToUnit{32.6pt}){\color{black}\rule{\textwidth}{.4pt}}%
  }%
}
\raggedbottom
\brokenpenalty=100
\clubpenalty=10000
\widowpenalty=10000
\maketitle
\renewcommand{\endabstract}{\par}
\patchcmd{\abstract}{\vspace*{12pt}}{\vspace*{7pt}}{}{}
\begin{abstract}
\looseness=1
A camouflaged object can blend into its surroundings at one scale and become distinguishable at another. Yet a region's apparent ambiguity gives limited guidance on which observation would resolve it. The challenge is to connect the form and scale of concealment to the evidence that a new view can reveal. We introduce \mbox{\method}, an active framework for camouflaged object detection guided by a Concealment Spectrum. From the image and current predictions, the spectrum measures regional resemblance and boundary ambiguity across comparison scales, distinguishing persistent concealment from separation emerging at finer or broader scales. These measurements guide crop selection from the already-captured image and recognition against global foreground and background references. An object-conditioned spatial preference then selects where local evidence should update the prediction and feature memory, retaining target corrections while suppressing background distractions. Accepted corrections reshape the next measurement, while predicted segmentation benefit determines whether to observe again. Experiments across encoder families and related segmentation tasks show consistent improvements. Targeted analyses support improved small-target recovery, multiple-object coverage, and \mbox{selective evidence integration. Code will be released.}
\end{abstract}

\newpage
\setlength{\abovedisplayskip}{2pt}
\setlength{\belowdisplayskip}{2pt}
\section{Introduction}
\label{sec:intro}
\begin{figure*}[t]
\centering
\includegraphics[width=\textwidth]{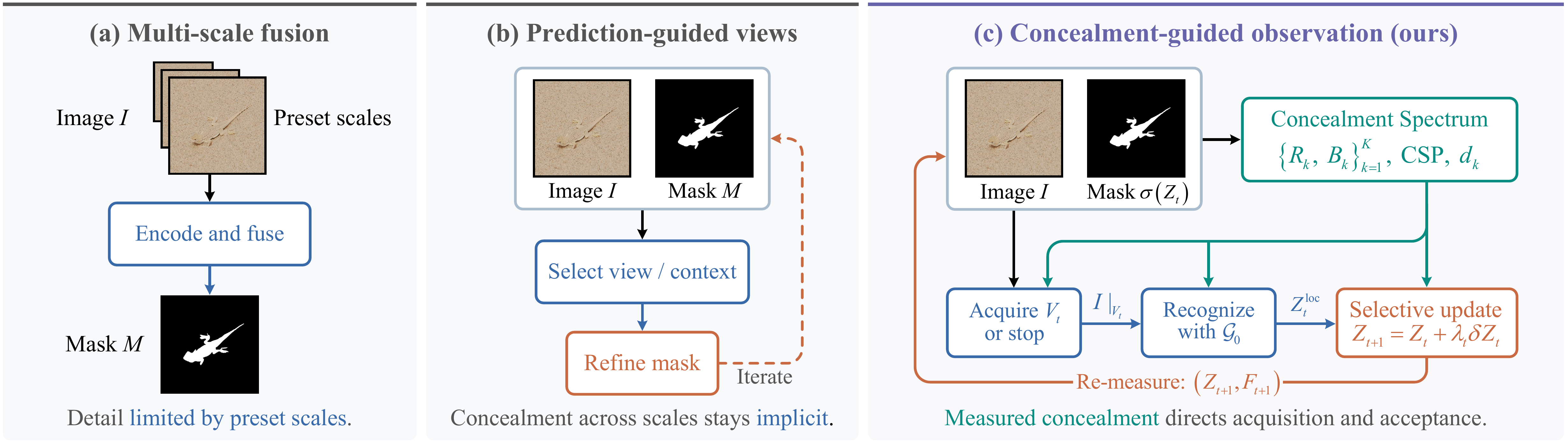}
\caption{\textbf{Observation strategies for COD.} Shared concealment evidence connects the choice of a view to the acceptance of its prediction.}
\label{fig:framework_comparison}
\end{figure*}

Camouflage succeeds by making an object resemble its surroundings. Yet the resemblance that hides it can change with the scale of comparison. In Fig.~\ref{fig:teaser}, a spider may blend into a curled leaf while retaining delicate leg contours; a seahorse may resemble nearby coral while differing from its wider surroundings; a moth may match coarse bark patterns while retaining a faint wing boundary at finer scales. For camouflaged object detection (COD)~\cite{le2019camo,fan2020sinet,fan2022concealed}, the question is: \emph{at which scales does camouflage break?}

A recognizer with a fixed input resolution must balance target detail and background context. Tightening a view allocates more pixels to the object; widening it preserves more of the surroundings. Low prediction confidence alone does not specify which change is useful. A barely visible contour may benefit from finer detail, whereas a region resembling its immediate neighborhood may require a broader comparison. Observation should therefore depend on how the object is concealed.

Lamdouar et al.~\cite{lamdouar2023making} quantify foreground--background similarity and boundary visibility to assess and synthesize camouflage. ZoomNet and ZoomNeXt aggregate preset image scales~\cite{pang2022zoomnet,pang2024zoomnext} (Fig.~\ref{fig:framework_comparison}a). SegMaR progressively magnifies predicted regions~\cite{jia2022segmar}; AdaptCOD adapts contextual support and retrieves background cues~\cite{yin2025adaptcod} (Fig.~\ref{fig:framework_comparison}b). We connect these directions by using scale-dependent concealment evidence as an input to observation decisions. A region's resemblance to its surroundings, and how that resemblance changes with comparison scale, can inform both the view to acquire and the local evidence to retain.

We introduce \method, a concealment-guided observation framework (Fig.~\ref{fig:framework_comparison}c). Its \emph{Concealment Spectrum} measures regional resemblance and boundary ambiguity across comparison scales using the image and current predictions. Concealment-scale persistence (CSP) summarizes sustained concealment; signed responses retain whether separation emerges toward finer or broader comparisons (Fig.~\ref{fig:evidence}). Conditioned on this profile, a learned policy chooses a native-image window's center and extent and predicts whether another observation is worthwhile.

Acquisition and integration share the same concealment evidence. The recognizer interprets local detail against global foreground and background references. A view can recover a target boundary while also strengthening a nearby distractor; an object-conditioned spatial preference therefore selects where to update segmentation and feature memory (Fig.~\ref{fig:trace}). Accepted corrections revise the supports used for subsequent measurement, linking an observation's effect to the next choice of where and how to observe.

Our contributions are: (1) an object-relative Concealment Spectrum that retains the strength and scale dependence of regional and boundary concealment; (2) spectrum-guided acquisition and recognition of native-image views; and (3) object-conditioned integration that closes the measurement--observation loop through selective updates. Window and integration controls support both observation selection and selective acceptance.

\flushbottom
\section{Related Work}
\label{sec:related}
\noindent\textbf{\mbox{Camouflaged object detection.}} COD methods strengthen prediction through search and segmentation~\cite{fan2020sinet,fan2022concealed}, high-resolution feedback~\cite{hu2023hitnet}, transformer aggregation~\cite{huang2023fspnet,yin2024camoformer}, frequency and boundary cues~\cite{he2023feder,sun2024fsel,escnet2025}, diffusion~\cite{sun2025camodiff,fu2026codiff}, and foundation representations~\cite{jan2025c3net,depthsam2026}. Our focus is how scale-dependent concealment can guide the evidence supplied to these representations.

\noindent\textbf{\mbox{Measuring concealment.}} Camouflage assessment studies background matching and edge visibility~\cite{troscianko2017quantifying}. Lamdouar et al.~\cite{lamdouar2023making} quantify foreground--background similarity and boundary visibility for dataset assessment and camouflage synthesis; RankNet estimates relative camouflage difficulty~\cite{lv2021ranknet}. Computed from the input image and predicted object masks, our spectrum tracks target--background similarity and boundary visibility across scales. It guides crop selection and local prediction fusion.

\noindent\textbf{\mbox{Adaptive observation and refinement.}} Dynamic Zoom-In Network selects regions using predicted detection gains and computation cost~\cite{gao2018dynamic}. MagNet progressively refines segmentation across magnification levels at uncertain locations~\cite{huynh2021progressive}. In COD, ZoomNet and ZoomNeXt fuse preset scales~\cite{pang2022zoomnet,pang2024zoomnext}; SegMaR uses attention-based sampling for iterative magnification~\cite{jia2022segmar}. AdaptCOD adjusts crop context using predicted foreground occupancy and retrieves global background representations~\cite{yin2025adaptcod}. \mbox{\method} conditions window geometry, recognition, and spatial acceptance on measured concealment responses. These describe object--surrounding relations across scales; learned windows balance target detail and background context.

\begingroup
\newcommand{\methodmathspacing}{%
  \setlength{\abovedisplayskip}{4pt plus 1pt minus 1pt}%
  \setlength{\belowdisplayskip}{4pt plus 1pt minus 1pt}%
  \setlength{\abovedisplayshortskip}{2pt plus 1pt}%
  \setlength{\belowdisplayshortskip}{4pt plus 1pt minus 1pt}}
\AtBeginEnvironment{equation}{\methodmathspacing}
\AtBeginEnvironment{align}{\methodmathspacing}

\begin{figure*}[!t]
\centering
\includegraphics[width=\textwidth]{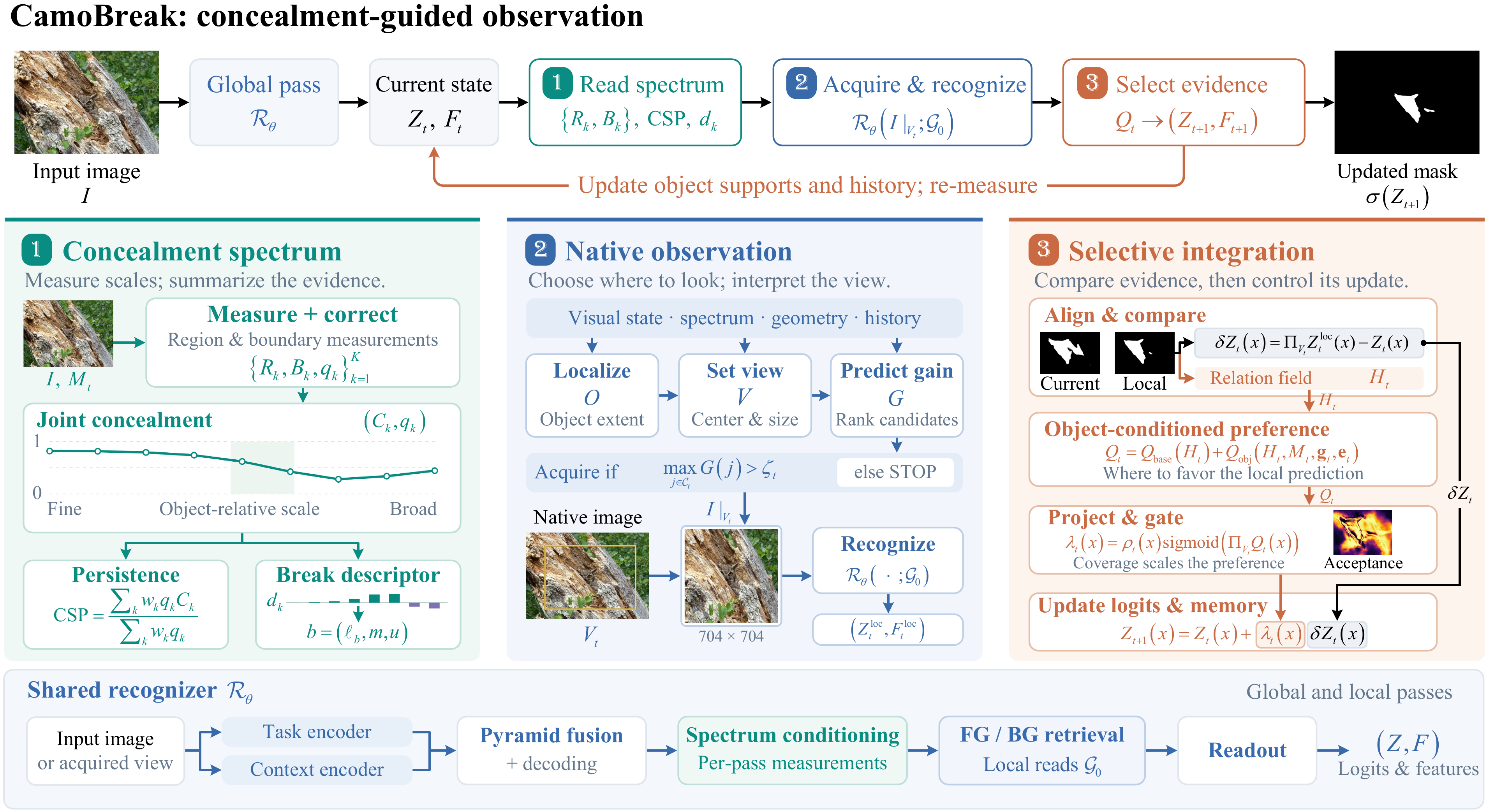}
\caption{\textbf{Schematic architecture of \method.} The Concealment Spectrum guides view selection and recognition. Object-conditioned preference integrates local evidence into predictions and feature memory; updated object supports define the next spectrum.}
\label{fig:loop}
\end{figure*}

\section{Method}
\label{sec:method}
\subsection{Overview}
For image $I\in\R^{H\times W\times3}$, a preliminary global mask supplies supports for measuring regional and boundary concealment across scales (Sec.~\ref{sec:measurement}). The shared recognizer $\mathcal R_\theta$ uses this spectrum to initialize native-grid logits $Z_0$, feature memory $F_0$, and global references $\mathcal G_0$ (Fig.~\ref{fig:loop}). The spectrum guides localization $O$, view selection $V$, and benefit prediction $G$; selected crops from $I$ are recognized with context from $\mathcal G_0$ (Sec.~\ref{sec:observe}). Object-conditioned preference $Q$ selects where local evidence updates $(Z_t,F_t)$ (Sec.~\ref{sec:writeback}). Updated supports define the next spectrum. Observation stops at the cap or when no eligible view has positive calibrated benefit. Training proceeds in stages (Sec.~\ref{sec:learning}).

\subsection{Concealment and Break across Scales}
\label{sec:measurement}
\paragraph{Object-relative measurement.}
For a predicted foreground support $M$ and the union $U$ of all supports, the background reference excludes $U$. We measure at $K$ logarithmically spaced scales $s_k\in[s_{\min},s_{\max}]$, with spatial support $h_k=s_kD$ and equivalent diameter $D=2\sqrt{|M|/\pi}$. Object size sets the reference unit $D$. Measurement scale $s_k$ varies the comparison from local detail to broader appearance and is distinct from the magnification of an acquired view. Both descriptor bandwidth and background-ring width vary with $s_k$, while the target support stays fixed. Changes along the profile therefore reflect the comparison scale, rather than a changing set of foreground pixels.

\paragraph{Region matching.}
A fixed descriptor encodes color, gradient magnitude, and local variation. Weighted target and background-ring moments yield $(\mu_k^+,\Sigma_k^+)$ and $(\mu_k^-,\Sigma_k^-)$. Their Gaussian affinity is
\begin{equation}
\begin{aligned}
 R_k&=\exp\!\left(-D_k^{\mu}-D_k^{\Sigma}\right),\\
 D_k^{\mu}&=\tfrac18\Delta\mu_k^\top\bar\Sigma_k^{-1}\Delta\mu_k,\\
 D_k^{\Sigma}&=\tfrac12\log
 \frac{\det\bar\Sigma_k}{\sqrt{\det\Sigma_k^+\det\Sigma_k^-}}.
\end{aligned}
\label{eq:region}
\end{equation}
Here $\bar\Sigma_k=(\Sigma_k^++\Sigma_k^-)/2$ and $\Delta\mu_k=\mu_k^+-\mu_k^-$. Moments of the fixed descriptor use all supported samples; high affinity indicates strong regional concealment.

\paragraph{Boundary concealment.}
Similar regions may still have a visible contour. At each foreground--background interface location $p$, Gaussian-weighted statistics give mean color vectors $\mu_k^\pm(p)$ and within-side deviations $\sigma_k^\pm(p)$. We define
\begin{align}
 v_k(p)&=\operatorname{clip}_{[0,1]}
 \frac{\|\mu^+_k(p)-\mu^-_k(p)\|_2}
 {\tau_{\rm col}\!\left[\sigma^+_k(p)+\sigma^-_k(p)+\epsilon_B\right]},\label{eq:visibility}\\
 B_k&=1-\frac{\sum_p e_k(p)v_k(p)}{\sum_p e_k(p)},\label{eq:boundary}
\end{align}
Here $\tau_{\rm col}>0$ normalizes contrast, $\epsilon_B$ stabilizes the ratio, and $e_k$ weights valid interfaces, excluding crop edges and other foreground objects. Weak contrast yields high $B_k$.

\paragraph{Persistence and break.}
The joint profile combines regional resemblance and boundary concealment, with reliability from both axes:
\begin{equation}
 C_k=\left(\frac{R_k^{\kappa}+B_k^{\kappa}}{2}\right)^{1/\kappa},
 \qquad q_k=\sqrt{q^R_kq^B_k}.
 \label{eq:joint_profile}
\end{equation}
With $\kappa<0$, the less concealed axis dominates; $C_k=0$ by continuity when either axis is zero. High $C_k$ thus requires both forms of concealment. A reliability-weighted integral summarizes persistence on the logarithmic grid:
\begin{equation}
 \CSP(M,I)=\frac{\sum_{k=1}^{K}w_kq_k C_k}{\sum_{k=1}^{K}w_kq_k}.
 \label{eq:csp}
\end{equation}
The trapezoidal weights are $w=(\tfrac12,1,\ldots,1,\tfrac12)$. Similar CSP values can hide opposite scale responses. We retain their direction through the signed change
\begin{equation}
 d_k=\frac{C_k-C_{k+1}}{C_k+C_{k+1}+\epsilon_\Delta}\min(q_k,q_{k+1}).
 \label{eq:scale_change}
\end{equation}
Positive $d_k$ marks separation under broader comparison; negative $d_k$ retains a distinction visible at finer scale. Break strength $m$, location $\ell_b$, and unresolved coarse-end concealment $u$ summarize the profile:
\begin{equation}
 m=\sum_{k=1}^{K-1}|d_k|,\quad
 \ell_b=\frac{\sum_{k=1}^{K-1}\tau_k|d_k|}{\max(m,\epsilon)},\quad
 u=q_KC_K,
 \label{eq:break}
\end{equation}
where $\tau_k=(k-\tfrac12)/(K-1)$ is the normalized log-scale midpoint. A flat profile can indicate either consistent separation or persistent concealment, which CSP distinguishes through its level. The descriptor $\boldsymbol b=(\ell_b,m,u)$ summarizes responsive scales and remaining concealment (Fig.~\ref{fig:evidence}).

\begin{figure}[!t]
\centering
\includegraphics[width=\columnwidth]{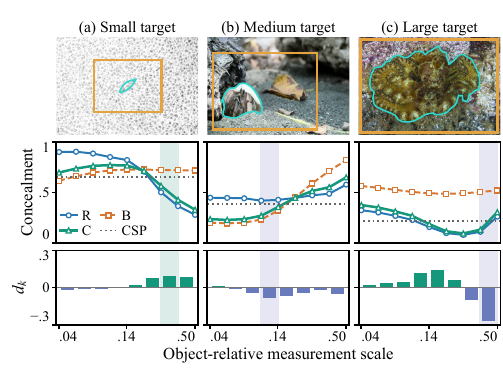}
\caption{\textbf{Concealment and separation across scales.} Distinct scale responses motivate object-specific observations. Shading marks the largest absolute signed response.}
\label{fig:evidence}
\end{figure}

\begin{figure*}[!t]
\centering
\includegraphics[width=\textwidth]{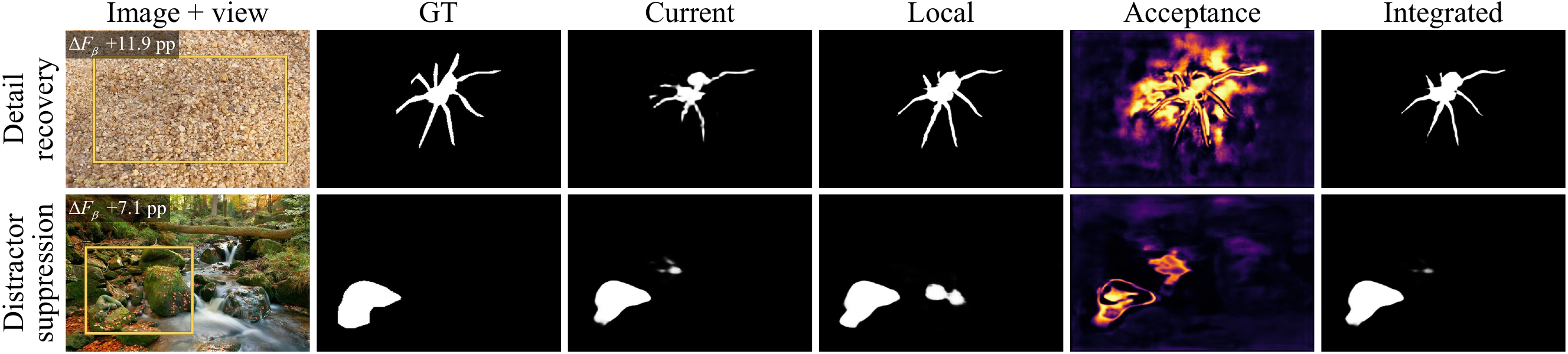}
\caption{\textbf{Selective integration of local evidence.} Spatial acceptance preserves target contours while rejecting background.}
\label{fig:trace}
\end{figure*}

\paragraph{Supported evidence.}
Reliability reflects native sample mass, resolvable scale, and visible interface support. A frozen probe estimates $q^R_k,q^B_k\in[0,1]$. Recognition receives neutralized fields $c_{\varnothing}+q^a_k(a_k-c_{\varnothing})$, $a\in\{R,B\}$, with $c_{\varnothing}=.5$, shrinking unreliable measurements toward a neutral value. Without reliable scales, CSP takes $c_{\varnothing}$.

\paragraph{Measurement with learned correction.}
To limit repeated measurement cost, $R/B$ are computed on an $L_m\times L_m$ grid around each predicted support. An MLP $f$ corrects resampling effects using formula values $\boldsymbol c_{L_m}$, probe curve estimates $\boldsymbol c_{\rm probe}$, and support geometry $\boldsymbol g$:
\begin{equation}
\begin{aligned}
 \boldsymbol\delta_c&=\alpha_c\tanh
 f(\boldsymbol c_{L_m},\boldsymbol g,\boldsymbol c_{\rm probe}),\\
 \hat{\boldsymbol c}&=\operatorname{clip}_{[0,1]}
 \left(\boldsymbol c_{L_m}+\boldsymbol\delta_c\right).
\end{aligned}
\end{equation}
Here $\alpha_c$ bounds the correction. Training targets are formula measurements on predicted supports from training images, with patch aspect ratio preserved. Inference requires only the image and predicted supports. Corrected evidence and probe reliability are shared by downstream decisions.

\subsection{Acquiring Revealing Observations}
\label{sec:observe}
\paragraph{From measured break to feature separation.}
The recognizer fuses task and context encoders with independent weights. Region, boundary, and CSP fields modulate decoded features $F^c$, followed by break conditioning:
\begin{equation}
 F^b=F^c\odot\big[1+\tanh\gamma_b(\boldsymbol b)\big]+\beta_b(\boldsymbol b).
 \label{eq:break_conditioning}
\end{equation}
Here $\gamma_b,\beta_b$ scale and shift features over each support. Conditioned global features, preliminary foreground probabilities, and spectrum descriptors form $\mathcal G_0$. Attention uses visual and spectral similarity, with foreground/background probabilities weighting separate retrievals whose difference refines recognition. Local outputs $(Z_t^{\rm loc},F_t^{\rm loc})$ combine detail from the current view with context retrieved from $\mathcal G_0$.

\paragraph{From scale response to a view.}
A learned residual refines current logits from features and measurements; thresholded foreground components form supports $M$. Overlap matching preserves observation history through splits and merges. Because these supports can be incomplete, localization $O$ combines their features with global context to estimate the target's full extent for window selection.

The view head pools features from this extent and four directional context regions, combining geometry and history with profiles, CSP, break strength, moment $\sum_{k=1}^{K-1}\tau_k|d_k|$, and coarse-end concealment. Reliability scales the spectral contribution. A shared offset $\eta_v$ calibrates four directional outputs: opposite-side differences adjust position parameters, and their sums adjust size parameters. These updates modify the localized box through a bounded transform, giving an in-image window for differentiable native-image sampling~\cite{jaderberg2015stn}.

For window area $|V_t|$ and input size $L\times L$, sampling density is $L^2/|V_t|$. Scale response, object extent, and observation history guide the balance between detail and context; measurement scale does not directly prescribe crop size.

\paragraph{Expected benefit of observation.}
Scale response identifies potential separation; $G$ predicts the error reduction from observing it. Each training rollout supplies the target
\begin{equation}
 \Delta_t=\mathcal E(Z_t,Y)-\mathcal E(Z_{t+1},Y),
 \label{eq:observation_benefit}
\end{equation}
where $\mathcal E$ combines weighted segmentation error~\cite{margolin2014wfmeasure} and cross-entropy at native resolution, and $Y$ is the training label. Profiles, CSP, break descriptors, geometry, and history predict $\Delta_t$ before acquisition. The candidate set $\mathcal C_t$ contains one valid window per active track with observation budget remaining. The next action is
\begin{equation}
 j_t=\begin{cases}
 \argmax_{j\in\mathcal C_t}G(j),&\max_{j\in\mathcal C_t}G(j)>\zeta_t,\\
 \mathrm{STOP},&\text{otherwise}.
 \end{cases}
 \label{eq:stopping}
\end{equation}
The stage threshold $\zeta_t$ calibrates benefit as $G(j)-\zeta_t$ and is shared across datasets and backbones. Observation also stops at cap $T_{\max}$ or with no candidates. Further observation thus requires predicted segmentation benefit.

\subsection{Aggregating Evidence under Concealment}
\label{sec:writeback}
A local view may reveal the target and a distractor together. Aligned features, signed and absolute feature differences, both predictions, projected spectrum fields, and coordinates form a relation field $H_t$. An object-conditioned residual augments shared spatial preference:
\begin{equation}
 Q_t=Q_{\rm base}(H_t)+Q_{\rm obj}(H_t,M_t,\boldsymbol g_t,\boldsymbol e_t).
 \label{eq:preference_field}
\end{equation}
Here $\boldsymbol g_t$ encodes target-relative coordinates, window ratios, and history. Evidence $\boldsymbol e_t$ retains profiles, signed changes, CSP, break strength and unnormalized moment, and \mbox{coarse-end} concealment; recognition and gain use normalized \mbox{location} $\ell_b$. Support $M_t$ identifies the target whose prediction the local evidence is intended to refine.

With native-grid projection $\Pi_{V_t}$ and fractional pixel coverage $\rho_t(x)\in[0,1]$, we integrate local evidence as
\begin{align}
 \lambda_t(x)&=\rho_t(x)\operatorname{sigmoid}\!\left(\Pi_{V_t}Q_t(x)\right),\\
 \delta Z_t(x)&=\Pi_{V_t}Z_t^{\rm loc}(x)-Z_t(x),\notag\\
 Z_{t+1}(x)&=Z_t(x)+\lambda_t(x)\,\delta Z_t(x).
 \label{eq:writeback}
\end{align}
$Q_t$ is projected to each memory grid for fusion. Since $\lambda_t\in[0,1]$, integration interpolates between the two logits, preserving unobserved pixels and leaving agreement unchanged. Spatially varying acceptance retains target corrections while suppressing background distractions (Fig.~\ref{fig:trace}). This decision also shapes subsequent observations: revised supports determine what is measured next, and updated feature memory guides view selection.

\subsection{Learning to Acquire and Aggregate}
\label{sec:learning}
Training fits the formula-supervised probe, recognizer, and measurement correction in order. Subsequent rollouts train localization, view selection, preference, and gain with fixed measurement and states detached between observations. Pixelwise and object/region structure losses supervise global, local, and integrated predictions. View supervision uses complete matched target components on a fixed domain, so the training loss accounts for all target pixels, including foreground outside the selected crop.

For current/local probabilities $p_t,p_t^{\rm loc}$ and label $y\in\{0,1\}$, preference supervision uses the signed difference
\begin{equation}
 (2y-1)(p_t^{\rm loc}-p_t)
 \label{eq:preference}
\end{equation}
whose sign identifies the prediction closer to the label. Disagreement magnitude and valid coverage weight binary cross-entropy, with balanced foreground/background contributions. A local increase is rewarded for recovering foreground and penalized for introducing a distractor. The gain head regresses $\Delta_t$ in Eq.~\ref{eq:observation_benefit}; at inference, predicted benefit determines whether to acquire another view.
\par\endgroup

\flushcolsend
\setlength{\floatsep}{6pt plus 1pt minus 1pt}
\setlength{\textfloatsep}{8pt plus 1pt minus 1pt}
\setlength{\intextsep}{4pt plus 1pt minus 1pt}
\setlength{\dblfloatsep}{6pt plus 1pt minus 1pt}
\setlength{\dbltextfloatsep}{8pt plus 1pt minus 1pt}
\begin{table*}[!t]
\begin{minipage}{\textwidth}
\captionsetup{type=table}
\centering
\caption{\textbf{COD comparison by backbone family.} \bestscore{Red}/\secondscore{blue}: best/second best within each family.}
\label{tab:context}
\compacttable\setlength{\tabcolsep}{3.4pt}
\setlength{\arrayrulewidth}{.3pt}
\fullwidthtable{34}{%
\begin{tabular}{l|cccc|cccc|cccc|cccc}
\toprule
&\multicolumn{4}{c|}{\textit{CHAMELEON}}&\multicolumn{4}{c|}{\textit{CAMO}}&\multicolumn{4}{c|}{\textit{COD10K}}&\multicolumn{4}{c}{\textit{NC4K}}\\
\cline{2-17}
\rowcolor{tablehead}\cellcolor{white}Methods&$M\downarrow$&$F_\beta\uparrow$&$E_\phi\uparrow$&$S_\alpha\uparrow$&$M\downarrow$&$F_\beta\uparrow$&$E_\phi\uparrow$&$S_\alpha\uparrow$&$M\downarrow$&$F_\beta\uparrow$&$E_\phi\uparrow$&$S_\alpha\uparrow$&$M\downarrow$&$F_\beta\uparrow$&$E_\phi\uparrow$&$S_\alpha\uparrow$\\
\midrule
\multicolumn{17}{c}{\textit{ResNet-50}~\cite{he2016resnet}}\\[-1pt]
SINet~\cite{fan2020sinet}&.034&.823&.936&.872&.092&.712&.804&.745&.043&.667&.864&.776&.058&.768&.871&.808\\
FSEL~\cite{sun2024fsel}&.027&.849&.946&.898&\secondscore{.067}&.801&.883&.823&.031&.742&.899&.832&\secondscore{.042}&.818&.907&.855\\
ZoomNeXt~\cite{pang2024zoomnext}&\bestscore{.021}&\secondscore{.873}&.942&\secondscore{.908}&\bestscore{.065}&\secondscore{.819}&.875&\secondscore{.833}&\secondscore{.026}&\secondscore{.769}&.906&\secondscore{.861}&\bestscore{.037}&\secondscore{.839}&\secondscore{.913}&\bestscore{.874}\\
RUN~\cite{he2025run}&.027&.855&.952&.895&.070&.781&.868&.806&.030&.747&.903&.827&\secondscore{.042}&.824&.908&.851\\
NUN~\cite{he2026nun}&.026&.862&\secondscore{.955}&.899&.069&.788&\bestscore{.890}&.818&.029&.759&\secondscore{.908}&.831&\secondscore{.042}&.828&\secondscore{.913}&.855\\
\rowcolor{tableours}\bfseries \method&\secondscore{.023}&\bestscore{.899}&\bestscore{.962}&\bestscore{.921}&\secondscore{.067}&\bestscore{.823}&\secondscore{.889}&\bestscore{.834}&\bestscore{.024}&\bestscore{.814}&\bestscore{.925}&\bestscore{.868}&\bestscore{.037}&\bestscore{.846}&\bestscore{.919}&\secondscore{.872}\\
\midrule
\multicolumn{17}{c}{\textit{Res2Net-50}~\cite{gao2021res2net}}\\[-1pt]
FSEL~\cite{sun2024fsel}&.026&.847&.951&.901&\secondscore{.065}&.803&.893&.832&.030&.739&.909&.839&.042&.820&.913&.861\\
RUN~\cite{he2025run}&.024&.879&.956&.907&.066&.815&.905&.843&.028&.764&.914&.849&.041&.830&.917&.859\\
NUN~\cite{he2026nun}&\secondscore{.023}&\secondscore{.883}&\secondscore{.959}&\secondscore{.910}&.066&\secondscore{.822}&\secondscore{.909}&\secondscore{.846}&\secondscore{.026}&\secondscore{.781}&\secondscore{.920}&\secondscore{.852}&\secondscore{.040}&\secondscore{.836}&\secondscore{.922}&\secondscore{.864}\\
\rowcolor{tableours}\bfseries \method&\bestscore{.018}&\bestscore{.898}&\bestscore{.966}&\bestscore{.921}&\bestscore{.049}&\bestscore{.856}&\bestscore{.916}&\bestscore{.861}&\bestscore{.020}&\bestscore{.837}&\bestscore{.940}&\bestscore{.883}&\bestscore{.031}&\bestscore{.870}&\bestscore{.933}&\bestscore{.888}\\
\midrule
\multicolumn{17}{c}{\textit{PVTv2}~\cite{wang2022pvtv2}}\\[-1pt]
ZoomNeXt~\cite{pang2024zoomnext}&\bestscore{.016}&\secondscore{.906}&.961&\secondscore{.925}&\secondscore{.040}&.866&.937&\secondscore{.888}&\secondscore{.017}&.830&.945&\secondscore{.898}&.028&.875&.940&\secondscore{.900}\\
FSEL~\cite{sun2024fsel}&.022&.873&.958&.916&\secondscore{.040}&.864&\secondscore{.942}&.884&.021&.796&.937&.876&.030&.863&.942&.893\\
CamoDiffusion~\cite{sun2025camodiff}&.022&.868&.952&.908&.042&.853&.936&.878&.019&.815&.943&.883&.028&.858&.942&.895\\
ESCNet~\cite{escnet2025}&.022&.886&.959&.906&.041&\secondscore{.868}&.941&.874&.020&.820&.941&.873&.029&.875&.941&.890\\
RUN~\cite{he2025run}&.021&.877&.958&.916&.045&.861&.934&.877&.021&.810&.941&.878&.030&.868&.940&.892\\
NUN~\cite{he2026nun}&.021&.888&\secondscore{.963}&.918&.042&\secondscore{.868}&.939&.885&.018&\secondscore{.839}&\secondscore{.948}&.890&\secondscore{.027}&\secondscore{.880}&\secondscore{.947}&.897\\
\rowcolor{tableours}\bfseries \method&\secondscore{.017}&\bestscore{.917}&\bestscore{.964}&\bestscore{.928}&\bestscore{.037}&\bestscore{.893}&\bestscore{.944}&\bestscore{.893}&\bestscore{.015}&\bestscore{.877}&\bestscore{.955}&\bestscore{.907}&\bestscore{.025}&\bestscore{.900}&\bestscore{.951}&\bestscore{.906}\\
\midrule
\multicolumn{17}{c}{\textit{DINOv2}~\cite{oquab2024dinov2}}\\[-1pt]
C3Net~\cite{jan2025c3net}&.016&.905&.968&.927&.031&.892&.953&.904&.016&.852&.961&.898&.024&.894&.958&.913\\
CODiff~\cite{fu2026codiff}&.017&.893&.959&.912&.031&.899&.952&.900&\secondscore{.014}&.861&.956&.899&.021&.901&.956&.912\\
DepthSAM~\cite{depthsam2026}&.016&.909&.972&\secondscore{.936}&\secondscore{.028}&.906&\secondscore{.958}&\secondscore{.919}&\secondscore{.014}&.866&.964&\secondscore{.920}&.021&.909&.961&\secondscore{.929}\\
NUN~\cite{he2026nun}&\secondscore{.015}&\secondscore{.918}&\secondscore{.975}&.931&\secondscore{.028}&\secondscore{.908}&\bestscore{.965}&.911&\secondscore{.014}&\secondscore{.880}&\secondscore{.972}&.905&\secondscore{.020}&\secondscore{.913}&\secondscore{.967}&.920\\
\rowcolor{tableours}\bfseries \method&\bestscore{.014}&\bestscore{.933}&\bestscore{.978}&\bestscore{.941}&\bestscore{.024}&\bestscore{.923}&\bestscore{.965}&\bestscore{.923}&\bestscore{.011}&\bestscore{.911}&\bestscore{.975}&\bestscore{.934}&\bestscore{.017}&\bestscore{.930}&\bestscore{.969}&\bestscore{.932}\\
\bottomrule
\end{tabular}}
\end{minipage}
\par\vspace{6pt}
\begin{minipage}{\textwidth}
\captionsetup{type=table}
\centering
\begin{minipage}[t]{.235\textwidth}
\centering\captionsetup{font=footnotesize,skip=3pt}

\caption{\textbf{PIS (ETIS).}}
\label{tab:pis}
\compacttable\fontsize{7}{8}\selectfont\setlength{\tabcolsep}{1.2pt}
\setlength{\arrayrulewidth}{.3pt}
\fittable{\linewidth}{%
\begin{tabular}{l|ccc}\toprule
\rowcolor{tablehead}\cellcolor{white}Methods&mDice$\uparrow$&mIoU$\uparrow$&$S_\alpha\uparrow$\\\midrule
CASCADE~\cite{rahman2023cascade}&.781&.706&.853\\
Polyp-PVT~\cite{dong2023polyppvt}&.787&.706&.871\\
LSSNet~\cite{wang2024lssnet}&.779&.701&.867\\
RUN~\cite{he2025run}&.788&.709&.878\\
NUN~\cite{he2026nun}&\secondscore{.791}&\secondscore{.716}&\secondscore{.882}\\
\rowcolor{tableours}\bfseries \method&\bestscore{.819}&\bestscore{.750}&\bestscore{.894}\\
\bottomrule\end{tabular}}

\end{minipage}\hfill
\begin{minipage}[t]{.235\textwidth}
\centering\captionsetup{font=footnotesize,skip=3pt}

\caption{\textbf{TOD (GDD).}}
\label{tab:tod}
\compacttable\fontsize{7}{8}\selectfont\setlength{\tabcolsep}{.2pt}
\setlength{\arrayrulewidth}{.3pt}
\fittable{\linewidth}{%
\begin{tabular}{l|ccc}\toprule
\rowcolor{tablehead}\cellcolor{white}Methods&mIoU$\uparrow$&$F_\beta^{\max}\uparrow$&$M\downarrow$\\\midrule
RFENet~\cite{fan2023rfenet}&.886&.938&.057\\
IEBAF~\cite{han2024iebaf}&.887&.944&.056\\
GhostingNet~\cite{yan2024ghostingnet}&.893&.943&.054\\
RUN~\cite{he2025run}&.900&.948&.051\\
NUN~\cite{he2026nun}&\secondscore{.903}&\bestscore{.956}&\secondscore{.048}\\
\rowcolor{tableours}\bfseries \method&\bestscore{.909}&\secondscore{.953}&\bestscore{.044}\\
\bottomrule\end{tabular}}

\end{minipage}\hfill
\begin{minipage}[t]{.275\textwidth}
\centering\captionsetup{font=footnotesize,skip=3pt}

\caption{\textbf{CDD (CDS2K).}}
\label{tab:cdd}
\compacttable\fontsize{7}{8}\selectfont\setlength{\tabcolsep}{1.5pt}
\setlength{\arrayrulewidth}{.3pt}
\fittable{\linewidth}{%
\begin{tabular}{l|cccc}\toprule
\rowcolor{tablehead}\cellcolor{white}Methods&$M\downarrow$&$F_\beta\uparrow$&$E_\phi\uparrow$&$S_\alpha\uparrow$\\\midrule
HitNet~\cite{hu2023hitnet}&.118&.298&.564&.563\\
FEDER~\cite{he2023feder}&.070&.288&.586&.538\\
CamoFormer~\cite{yin2024camoformer}&.100&\secondscore{.330}&.588&.590\\
RUN~\cite{he2025run}&\secondscore{.068}&.298&.595&.590\\
NUN~\cite{he2026nun}&\bestscore{.066}&.303&\secondscore{.599}&\secondscore{.595}\\
\rowcolor{tableours}\bfseries \method&.085&\bestscore{.375}&\bestscore{.621}&\bestscore{.607}\\
\bottomrule\end{tabular}}

\end{minipage}\hfill
\begin{minipage}[t]{.235\textwidth}
\centering\captionsetup{font=footnotesize,skip=3pt}

\caption{\textbf{MTOS (DRIVE).}}
\label{tab:mtos}
\compacttable\fontsize{7}{8}\selectfont\setlength{\tabcolsep}{1.5pt}
\setlength{\arrayrulewidth}{.3pt}
\fittable{\linewidth}{%
\begin{tabular}{l|ccc}\toprule
\rowcolor{tablehead}\cellcolor{white}Methods&mDice$\uparrow$&AUC$\uparrow$&SEN$\uparrow$\\\midrule
CS2-Net~\cite{mou2021cs2net}&.795&.983&.822\\
DSCNet~\cite{qi2023dscnet}&.805&.955&.830\\
SGAT~\cite{lin2023sgat}&.806&.953&.832\\
FFM~\cite{huang2024ffm}&.791&.972&.830\\
RUN~\cite{he2025run}&\secondscore{.812}&\secondscore{.985}&\secondscore{.845}\\
\rowcolor{tableours}\bfseries \method&\bestscore{.820}&\bestscore{.986}&\bestscore{.848}\\
\bottomrule\end{tabular}}

\end{minipage}

\end{minipage}
\par\vspace{6pt}
\begin{minipage}{\textwidth}
\centering
\caption{\textbf{Component ablations on COD.} Per-image averages across four benchmarks.}
\label{tab:ablation_full}
\compacttable\setlength{\tabcolsep}{3pt}
\columntable{10}{%
\begin{tabular}{l|ccccc|c|c|c|c}\toprule
&\multicolumn{5}{c|}{Concealment spectrum}&Observation&References&Integration&\cellcolor{tableours}Full\\
\cline{2-9}
\rowcolor{tablehead}Metric&w/o $R$&w/o $B$&w/o CSP&\shortstack{w/o signed\\response/break}&\shortstack{w/o entire\\spectrum}&\shortstack{w/o spectral\\view inputs}&w/o $\mathcal G_0$&w/o $Q_{\rm obj}$&\cellcolor{tableours}\method\\\midrule
$M\downarrow$&.025&.024&.024&.024&.024&.024&.024&.024&\cellcolor{tableours}\textbf{.022}\\
$F_\beta\uparrow$&.881&.877&.878&.878&.883&.876&.873&.876&\cellcolor{tableours}\textbf{.893}\\
$E_\phi\uparrow$&.946&.946&.948&.947&.948&.944&.946&.948&\cellcolor{tableours}\textbf{.952}\\
$S_\alpha\uparrow$&.897&.896&.899&.897&.899&.895&.894&.897&\cellcolor{tableours}\textbf{.906}\\
\bottomrule\end{tabular}}
\end{minipage}
\end{table*}
\section{Experiments}
\label{sec:experiments}

\subsection{Experimental Setup}
\paragraph{Datasets and metrics.}
For COD, we train on CAMO and COD10K and evaluate on CHAMELEON~\cite{skurowski2017chameleon}, CAMO~\cite{le2019camo}, COD10K~\cite{fan2020sinet}, and NC4K~\cite{lv2021ranknet}. Predictions are evaluated at native resolution using MAE ($M$), adaptive $F$-measure ($F_\beta$, $\beta^2=.3$)~\cite{achanta2009ft}, mean enhanced alignment ($E_\phi$)~\cite{fan2018emeasure}, and structure measure ($S_\alpha$)~\cite{fan2017smeasure}.

\paragraph{Implementation.}
Training uses four NVIDIA RTX PRO 6000 GPUs. The default PVTv2-B4 backbone~\cite{wang2022pvtv2} uses $352\times352$ global and $704\times704$ local inputs. Both views share the encoders and decoder, fixed at inference. Automatic stopping caps local views at three. The supplement presents implementation details and additional experiments.

\subsection{Comparison with State-of-the-Art Methods}
\label{sec:comparison}
\begin{figure*}[!t]
\centering
\includegraphics[width=\textwidth]{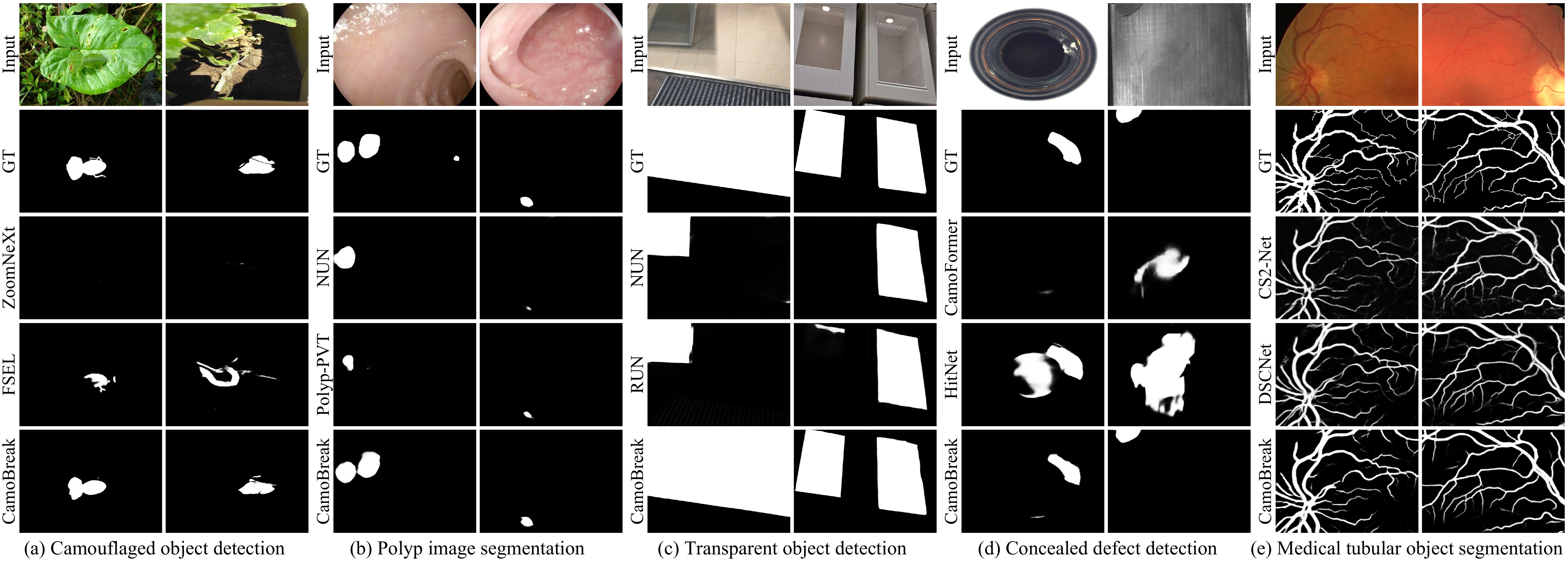}
\captionsetup{skip=2pt}
\caption{\textbf{Recovering concealed boundaries and thin structures across tasks.}}
\label{fig:qualitative}
\end{figure*}

\paragraph{Camouflaged object detection.}
Table~\ref{tab:context} compares convolutional, hierarchical-transformer, and foundation encoders. \method leads adaptive $F_\beta$ in all 16 backbone--dataset settings and $S_\alpha$ in 15. The default model improves COD10K $F_\beta$ from .839 for NUN to .877. These gains across encoder families support observation control alongside representation learning. Figure~\ref{fig:qualitative} illustrates the resulting recovery of target boundaries and thin structures.

\noindent\textbf{Generalization to concealed objects.}\quad
\label{sec:transfer}
We evaluate polyp image segmentation (PIS), transparent object \mbox{detection (TOD)}, concealed defect detection (CDD), and medical tubular object segmentation (MTOS) to test concealment-guided observation across weak boundaries, transparency, subtle defects, and thin structures.

\raggedbottom
PIS trains on Kvasir-SEG~\cite{jha2020kvasir} and CVC-ClinicDB~\cite{bernal2015clinicdb}, reporting threshold-averaged Dice and IoU on ETIS~\cite{silva2014etis} (Table~\ref{tab:pis}). TOD uses GDD~\cite{mei2020gdd}, threshold-0.5 IoU, and maximum $F_\beta$ from dataset-mean precision/recall (Table~\ref{tab:tod}). MTOS follows the official DRIVE split~\cite{staal2004drive} with per-image Dice, AUC, and sensitivity (Table~\ref{tab:mtos}). These tasks use separate training; CDD applies the COD model to positive CDS2K images~\cite{fan2023csu} without adaptation (Table~\ref{tab:cdd}).

\method leads the reported PIS and MTOS metrics, TOD IoU and MAE, and CDD $F_\beta$, $E_\phi$, and $S_\alpha$. On ETIS, mDice reaches .819 versus .791 for NUN. MTOS visualizations use our CS2-Net and DSCNet reproductions.

\Needspace{16\baselineskip}
\subsection{Ablation Study}
\label{sec:ablations}
\vspace{-5pt}
Each variant in Table~\ref{tab:ablation_full} is retrained independently. Spectrum removal disables measurement, correction, and spectral inputs to acquisition, recognition, and integration.

All deletions increase MAE and reduce $F_\beta$, $E_\phi$, and $S_\alpha$. Removing $R$ or $B$ retains the other multiscale profile, testing region and boundary evidence separately. CSP and signed-response removal retain both profiles but respectively discard aggregate concealment or explicit scale changes. Both give $F_\beta=.878$ versus .893 for the full model.
Global-reference removal yields the largest $F_\beta$ loss (.893 to .873), highlighting the context needed to interpret local detail. Spectral-view and object-preference removal each reach .876. The former keeps measurements for recognition and integration, so its loss supports their contribution to choosing observation windows.

\Needspace*{8\baselineskip}
\subsection{Further Analysis}
\label{sec:profile_analysis}
\raggedbottom
\noindent\textbf{Observation benefit and cost.}\quad
Local observation raises $F_\beta$ from .852 to .893 and lowers MAE from .027 to .022 (Table~\ref{tab:observation_resolution}), surpassing both baselines on all four metrics.

\par\addvspace{6pt}
\begin{nolinenumbers}
\noindent\begin{minipage}{\columnwidth}
\captionsetup{type=table}
\centering
\caption{\textbf{Local observation on COD.} {\method}$^\dagger$ uses \mbox{$352\times352$} local inputs.}
\label{tab:observation_resolution}
\compacttable
\columntable{5}{%
\begin{tabular}{l|cccc}\toprule
\rowcolor{tablehead}Methods&$M\downarrow$&$F_\beta\uparrow$&$E_\phi\uparrow$&$S_\alpha\uparrow$\\\midrule
ZoomNeXt~\cite{pang2024zoomnext}&\secondscore{.025}&.861&.942&\secondscore{.900}\\
FSEL~\cite{sun2024fsel}&.027&.842&.940&.888\\
\midrule
\rowcolor{tableours}\bfseries \method&\bestscore{.022}&\bestscore{.893}&\bestscore{.952}&\bestscore{.906}\\
\quad {\method}$^\dagger$&\secondscore{.025}&\secondscore{.875}&\secondscore{.946}&.895\\
\quad w/o local observations&.027&.852&.940&.885\\
\bottomrule\end{tabular}}
\end{minipage}
\end{nolinenumbers}
\par\addvspace{6pt}

At reduced local resolution, {\method}$^\dagger$ retains 58\% of the $F_\beta$ gain with one quarter of the local input pixels. It reaches $F_\beta=.875$ and $E_\phi=.946$, above .861 and .942 for ZoomNeXt. These gains at the global input size support the contribution of targeted observation and selective integration beyond increased resolution.

Figure~\ref{fig:budget} shows diminishing accuracy gains with more observations. Automatic stopping preserves three-observation accuracy and exceeds six-observation $F_\beta$ at lower latency. The model predicts benefit from the revised state after each update, so further observations focus on views that can still refine the current mask.

\par\addvspace{6pt}
\begin{nolinenumbers}
\noindent\begin{minipage}{\columnwidth}
\captionsetup{type=figure}
\centering
\includegraphics[width=\linewidth]{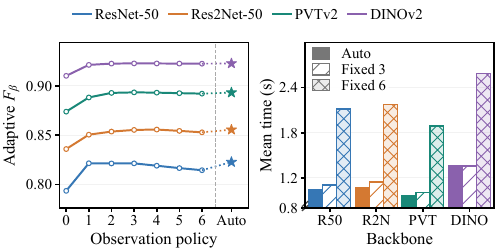}
\captionsetup{skip=2pt}
\caption{\textbf{Accuracy and observation cost on CAMO.} Time includes image loading, measurement, recognition, and integration.}
\label{fig:budget}
\end{minipage}
\end{nolinenumbers}
\par\addvspace{6pt}

\Needspace{18\baselineskip}
\noindent\textbf{Learning observation windows.}\quad
Table~\ref{tab:fixed_windows} compares windows after one local observation with frozen model weights. Fixed squares center on the current predicted support box, with sides $r$ times its longer side; recognition and integration are unchanged. Increasing the fixed ratio beyond $r=1.5$ does not improve $F_\beta$, despite including more context. Learned windows raise $F_\beta$ from .849 at the best fixed ratio ($r=1.5$) to .857, with higher $E_\phi$ and $S_\alpha$. The gain suggests that concealment cues help balance fine target detail with contextual evidence from the broader scene.

\par\addvspace{6pt}
\begin{nolinenumbers}
\noindent\begin{minipage}{\columnwidth}
\captionsetup{type=table}
\centering
\caption{\textbf{Fixed and learned windows on COD10K.} Windows are clipped to image boundaries.}
\label{tab:fixed_windows}
\compacttable
\columntable{5}{%
\begin{tabular}{l|cccc}\toprule
\rowcolor{tablehead}Window&$M\downarrow$&$F_\beta\uparrow$&$E_\phi\uparrow$&$S_\alpha\uparrow$\\\midrule
Fixed $r=1$&.018&.843&.946&.891\\
Fixed $r=1.5$&.017&.849&.948&.894\\
Fixed $r=2$&.018&.848&.948&.893\\
Fixed $r=3$&.018&.847&.948&.893\\
\midrule
\rowcolor{tableours}\bfseries Learned&\textbf{.017}&\textbf{.857}&\textbf{.950}&\textbf{.898}\\
\bottomrule\end{tabular}}
\end{minipage}
\end{nolinenumbers}
\par\addvspace{6pt}

\noindent\textbf{Selective local integration.}\quad
Table~\ref{tab:writeback} compares full replacement, uniform mixing, and spatial acceptance with matched incoming states and local predictions.
Selective integration more than doubles replacement's mean $F_\beta$ gain and reduces harmful updates from 45.4\% to 25.2\%, preserving target corrections while suppressing distractors (Fig.~\ref{fig:trace}).
Compared with equal mixing, it reduces harmful updates by 5.8 percentage points while improving average error reduction and $F_\beta$. The comparison supports choosing where local evidence should update the current prediction.

\par\addvspace{6pt}
\begin{nolinenumbers}
\noindent\begin{minipage}{\columnwidth}
\captionsetup{type=table}
\centering
\caption{\textbf{Integration on COD.} Means over 18,632 matched updates. For raw sigmoid probabilities $p$, $M_{\rm prob}=|\Omega|^{-1}\sum_{x\in\Omega}|p(x)-Y(x)|$ averages over all image pixels $\Omega$. An update is harmful when $\Delta M_{\rm prob}=M_{\rm prob}^{\rm before}-M_{\rm prob}^{\rm after}<0$.}
\label{tab:writeback}
\compacttable\setlength{\tabcolsep}{2.2pt}
\columntable{4}{%
\begin{tabular}{l|ccc}\toprule
\rowcolor{tablehead}Integration&\shortstack{$\Delta M_{\rm prob}\uparrow$\\($\times10^{-3}$)}&\shortstack{$\Delta F_\beta\uparrow$\\(pp)}&\shortstack{Harmful\\updates (\%)$\downarrow$}\\\midrule
Local replacement&+0.04&+0.60&45.4\\
Equal mixing&+1.42&+1.13&31.1\\
\midrule
\rowcolor{tableours}\bfseries \method&\textbf{+1.71}&\textbf{+1.42}&\textbf{25.2}\\
\bottomrule\end{tabular}}
\end{minipage}
\end{nolinenumbers}

\begin{table*}[!t]
\begin{minipage}[b]{\columnwidth}
\centering\captionsetup{type=table}
\begingroup
\caption{\textbf{Object-size analysis on COD10K.} $a$: instance area divided by image area.}
\label{tab:size}
\compacttable\setlength{\tabcolsep}{1.5pt}
\columntable{7}{%
\begin{tabular}{l|cc|cc|cc}\toprule
&\multicolumn{2}{c|}{\shortstack{Small\\$a<5\%$}}&\multicolumn{2}{c|}{\shortstack{Medium\\$5\%\leq a<25\%$}}&\multicolumn{2}{c}{\shortstack{Large\\$a\geq25\%$}}\\
&\multicolumn{2}{c|}{$n=918$}&\multicolumn{2}{c|}{$n=948$}&\multicolumn{2}{c}{$n=119$}\\
\rowcolor{tablehead}Methods&$F_\beta\uparrow$&$M\downarrow$&$F_\beta\uparrow$&$M\downarrow$&$F_\beta\uparrow$&$M\downarrow$\\\midrule
ZoomNeXt~\cite{pang2024zoomnext}&\secondscore{.738}&\secondscore{.010}&\secondscore{.908}&\secondscore{.019}&.895&\secondscore{.049}\\
FSEL~\cite{sun2024fsel}&.682&.014&.885&.023&\secondscore{.933}&.057\\
\midrule
\rowcolor{tableours}\bfseries \method&\bestscore{.817}&\bestscore{.008}&\bestscore{.924}&\bestscore{.018}&\bestscore{.947}&\bestscore{.046}\\
\quad w/o local observations&.705&.013&.889&.024&.932&.056\\
\bottomrule\end{tabular}}
\endgroup

\par\vspace{3pt}

\begin{captionblock}{\columnwidth}{figure}
\centering\captionsetup{skip=2pt}
\includegraphics[width=\columnwidth]{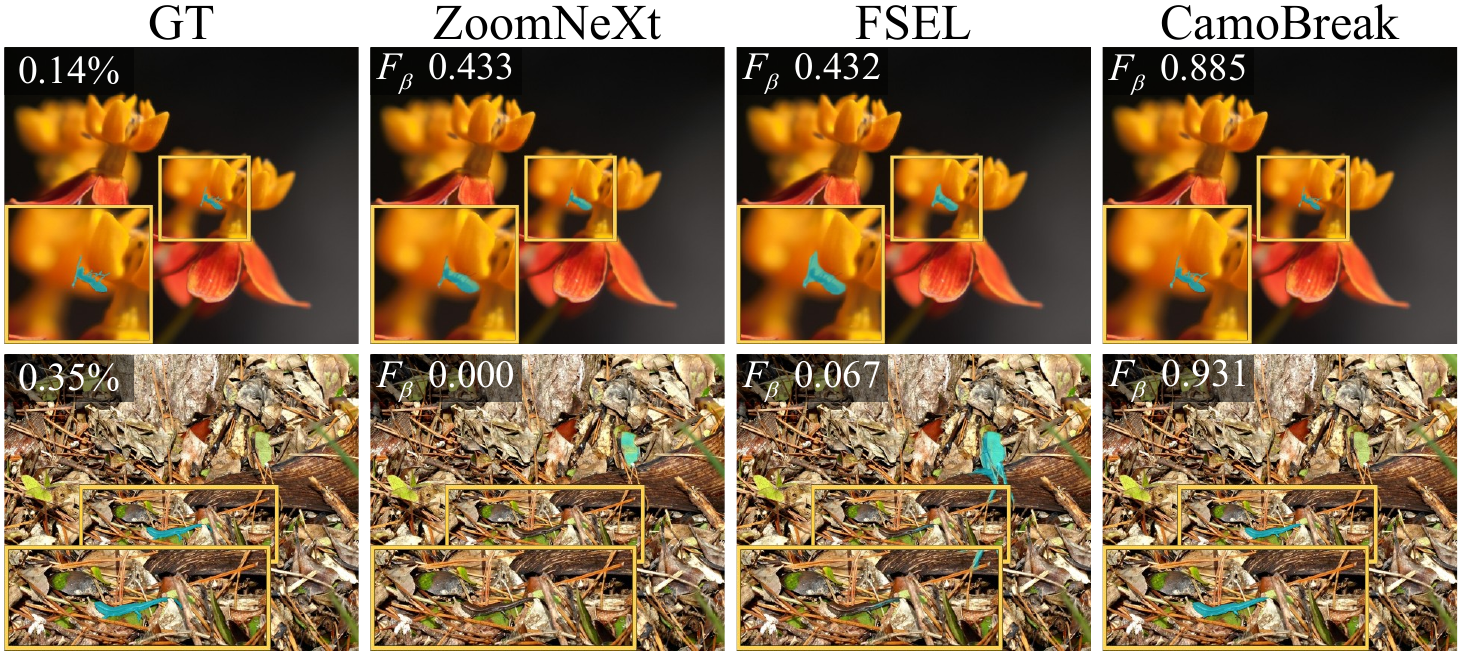}
\caption{\textbf{Recovering small targets on COD10K.} Yellow boxes link observation windows to enlarged views.}
\label{fig:small_objects}
\end{captionblock}

\end{minipage}\hfill
\begin{minipage}[b]{\columnwidth}
\centering\captionsetup{type=table}
\begingroup
\caption{\textbf{Multiple objects on NC4K.} All 330 images with at least two annotated objects.}
\label{tab:multiobject}
\compacttable\renewcommand{\arraystretch}{1.25}\setlength{\tabcolsep}{3pt}
\columntable{5}{%
\begin{tabular}{l|cccc}\toprule
\rowcolor{tablehead}Methods&$M\downarrow$&$F_\beta\uparrow$&$S_\alpha\uparrow$&Instance recall (\%)$\uparrow$\\\midrule
ZoomNeXt~\cite{pang2024zoomnext}&\secondscore{.040}&.856&\secondscore{.857}&\secondscore{74.7}\\
FSEL~\cite{sun2024fsel}&.046&.836&.841&73.7\\
\midrule
\rowcolor{tableours}\bfseries \method&\bestscore{.039}&\bestscore{.870}&\bestscore{.858}&\bestscore{74.8}\\
\quad w/o local observations&.047&.846&.838&71.7\\
\quad w/o object residual&.041&\secondscore{.868}&.853&73.6\\
\bottomrule\end{tabular}}
\endgroup

\par\vspace{3pt}

\begin{captionblock}{\columnwidth}{figure}
\centering\captionsetup{skip=2pt}
\includegraphics[width=\columnwidth]{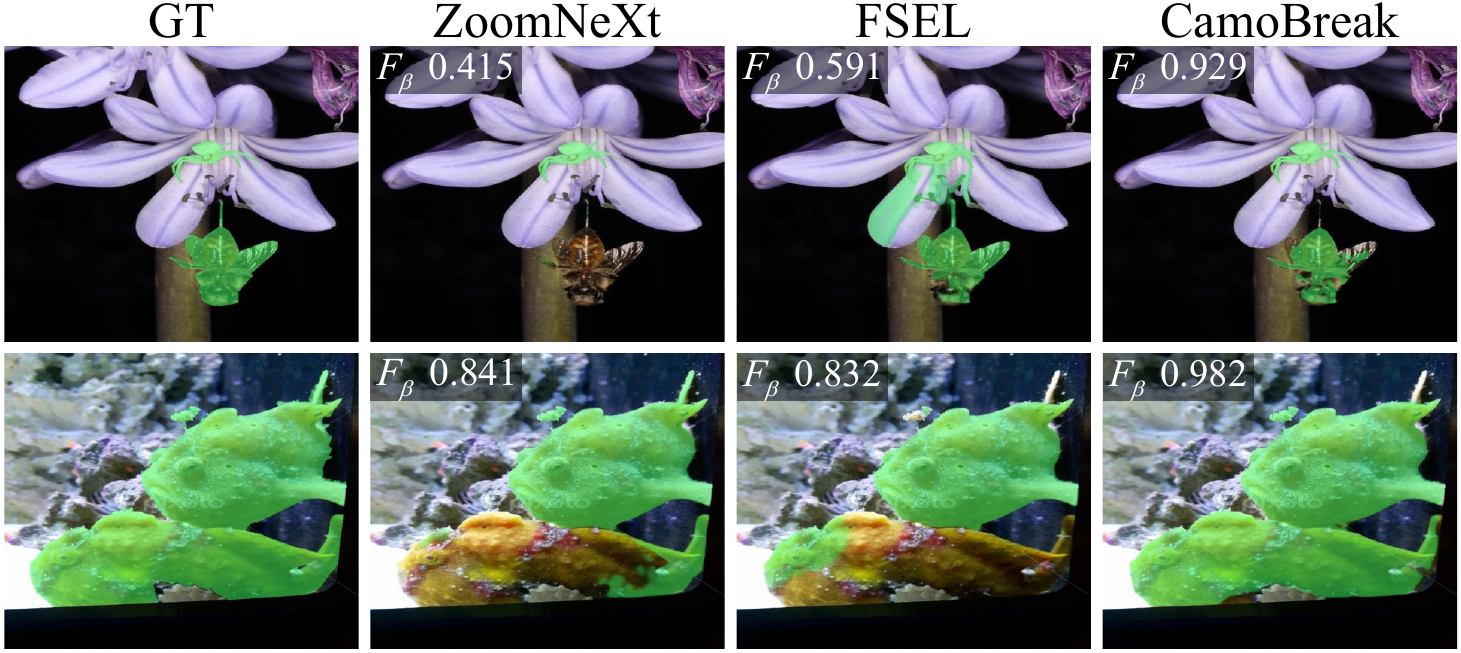}
\caption{\textbf{Preserving multiple targets on NC4K.} Object-conditioned integration supports coverage of distinct instances.}
\label{fig:multi_objects}
\end{captionblock}

\end{minipage}
\end{table*}
\par\addvspace{6pt}

\Needspace{12\baselineskip}
\paragraph{Recovering small targets.}
\label{sec:attributes}
Local observation yields its largest $F_\beta$ gain on small targets (Table~\ref{tab:size}). A COD10K image is assigned to a size group only if all its annotated instances fall within that interval. On small targets, observation raises $F_\beta$ from .705 to .817, compared with .738 for ZoomNeXt. This gain of 11.2 percentage points exceeds the 3.5 and 1.5 points for medium and large targets. The trend is consistent with native views recovering fine detail lost during global resizing (Fig.~\ref{fig:small_objects}).

\paragraph{Preserving multiple targets.}
Table~\ref{tab:multiobject} covers all NC4K images with multiple annotated objects. Instance recall uses a $0.5$ threshold after per-image min--max normalization, with equal averaging over instances within each image and then over images. Observation raises recall from 71.7\% to 74.8\%. Removing the object-conditioned residual lowers recall to 73.6\%, $F_\beta$ from .870 to .868, and $S_\alpha$ from .858 to .853. The object residual helps preserve multiple targets after local observation (Fig.~\ref{fig:multi_objects}).

\Needspace{5\baselineskip}
\section{Discussion}
\label{sec:discussion}
\method makes concealment a shared decision signal for acquiring, interpreting, and accepting evidence (Fig.~\ref{fig:framework_comparison}c). Preset-scale fusion (a) aggregates predefined views~\cite{pang2022zoomnet,pang2024zoomnext}, while prediction-guided observation (b) adapts magnification and context~\cite{jia2022segmar,yin2025adaptcod}. Our spectrum retains how regional resemblance and boundary ambiguity change with comparison scale, giving camouflage assessment~\cite{lamdouar2023making} a role in deciding how to observe. The policy learns window geometry from this profile and the evolving scene; comparison scale does not prescribe crop size. Learned windows outperform every tested fixed ratio with weights and \mbox{observation count} held fixed (Table~\ref{tab:fixed_windows}). Retrained ablations support direct spectral input to view selection and global references for recognition (Table~\ref{tab:ablation_full}). Acquisition alone is insufficient: a useful view can also strengthen a distractor. Beyond uncertainty-based refinement~\cite{huynh2021progressive}, our acceptance uses the selected object's support and concealment profile. It reduces harmful updates under matched incoming states and local predictions (Table~\ref{tab:writeback}). Accepted corrections revise the supports used for the next spectrum, so the effect of one observation changes the evidence for the next. This shared feedback connects the reason for seeking local detail or context to the decision to retain its contribution.

The loop depends on its current object supports: missed targets can remain unobserved, and similar distractors can contaminate measurements and acquired views. Selective integration reduces but does not eliminate harmful updates; partial coverage and foreground leakage can persist after observation. Native sampling cannot recover absent source detail, and repeated recognition incurs computation. Better candidate coverage and target discrimination remain complementary to observation control.

\section{Conclusion}
\method connects scale-dependent concealment to purposeful observation. Region and boundary profiles guide native views and reference-based recognition; object-conditioned preference retains useful evidence in prediction and feature memory. The resulting loop improves segmentation across encoder families and related tasks. On COD10K images with only small targets, our model reaches .817 adaptive $F_\beta$ versus .738 for \mbox{ZoomNeXt}.

\FloatBarrier
\flushcolsend
\clearpage
\raggedcolsend
\flushbottom
\makeatletter
\patchcmd{\set@outputbox@with@footnote@and@float}
  {\setbox\flushend@@tempbox@c\vsplit}
  {\@tempdimb=\@colht\setbox\flushend@@tempbox@c\vsplit}
  {}{\PackageError{camobreak}{Reference column alignment failed}{Check flushend compatibility.}}
\makeatother
\section*{LLM Usage Statement}
ChatGPT (OpenAI) was used to improve the language and clarity of the manuscript, review the code, and assist with the experiments. The authors take full responsibility for all content presented in this paper.

{\small
\makeatletter
\patchcmd{\cvprsection}{-10pt plus -2pt minus -2pt}{-5pt}{}{}
\makeatother
\setlength{\bibsep}{-.65pt plus .5pt}
\bibliographystyle{ieeenat_fullname}\bibliography{main}}

\begin{thebibliography}{50}
\providecommand{\natexlab}[1]{#1}
\providecommand{\url}[1]{\texttt{#1}}
\expandafter\ifx\csname urlstyle\endcsname\relax
  \providecommand{\doi}[1]{doi: #1}\else
  \providecommand{\doi}{doi: \begingroup \urlstyle{rm}\Url}\fi

\bibitem[Achanta et~al.(2009)Achanta, Hemami, Estrada, and S{\"u}sstrunk]{achanta2009ft}
Radhakrishna Achanta, Sheila Hemami, Francisco Estrada, and Sabine S{\"u}sstrunk.
\newblock Frequency-tuned salient region detection.
\newblock In \emph{IEEE Conference on Computer Vision and Pattern Recognition}, pages 1597--1604, 2009.

\bibitem[Bernal et~al.(2015)Bernal, S{\'a}nchez, Fern{\'a}ndez-Esparrach, Gil, Rodr{\'i}guez, and Vilari{\~n}o]{bernal2015clinicdb}
Jorge Bernal, F.~Javier S{\'a}nchez, Gloria Fern{\'a}ndez-Esparrach, Debora Gil, Cristina Rodr{\'i}guez, and Fernando Vilari{\~n}o.
\newblock {WM-DOVA} maps for accurate polyp highlighting in colonoscopy: Validation vs. saliency maps from physicians.
\newblock \emph{Computerized Medical Imaging and Graphics}, 43:\penalty0 99--111, 2015.

\bibitem[Dong et~al.(2023)Dong, Wang, Fan, Li, Fu, and Shao]{dong2023polyppvt}
Bo Dong, Wenhai Wang, Deng-Ping Fan, Jinpeng Li, Huazhu Fu, and Ling Shao.
\newblock {{Polyp-PVT}: Polyp Segmentation with Pyramid Vision Transformers}.
\newblock \emph{CAAI Artificial Intelligence Research}, 2:\penalty0 9150015, 2023.

\bibitem[Fan et~al.(2017)Fan, Cheng, Liu, Li, and Borji]{fan2017smeasure}
Deng-Ping Fan, Ming-Ming Cheng, Yun Liu, Tao Li, and Ali Borji.
\newblock {Structure-measure: A new way to evaluate foreground maps}.
\newblock In \emph{Int. Conf. Comput. Vis.}, pages 4548--4557, 2017.

\bibitem[Fan et~al.(2018)Fan, Gong, Cao, Ren, Cheng, and Borji]{fan2018emeasure}
Deng-Ping Fan, Cheng Gong, Yang Cao, Bo Ren, Ming-Ming Cheng, and Ali Borji.
\newblock {Enhanced-alignment measure for binary foreground map evaluation}.
\newblock In \emph{Int. Joint Conf. Artif. Intell.}, pages 698--704, 2018.

\bibitem[Fan et~al.(2020)Fan, Ji, Sun, Cheng, Shen, and Shao]{fan2020sinet}
Deng-Ping Fan, Ge-Peng Ji, Guolei Sun, Ming-Ming Cheng, Jianbing Shen, and Ling Shao.
\newblock {Camouflaged Object Detection}.
\newblock In \emph{IEEE/CVF Conference on Computer Vision and Pattern Recognition}, pages 2777--2787, 2020.

\bibitem[Fan et~al.(2022)Fan, Ji, Cheng, and Shao]{fan2022concealed}
Deng-Ping Fan, Ge-Peng Ji, Ming-Ming Cheng, and Ling Shao.
\newblock {Concealed object detection}.
\newblock \emph{IEEE Trans. Pattern Anal. Mach. Intell.}, 44\penalty0 (10):\penalty0 6024--6042, 2022.

\bibitem[Fan et~al.(2023{\natexlab{a}})Fan, Ji, Xu, Cheng, Sakaridis, and Van~Gool]{fan2023csu}
Deng-Ping Fan, Ge-Peng Ji, Peng Xu, Ming-Ming Cheng, Christos Sakaridis, and Luc Van~Gool.
\newblock {Advances in Deep Concealed Scene Understanding}.
\newblock \emph{Visual Intelligence}, 1:\penalty0 16, 2023{\natexlab{a}}.

\bibitem[Fan et~al.(2023{\natexlab{b}})Fan, Wang, Wang, Wang, Yi, and Ma]{fan2023rfenet}
Ke Fan, Changan Wang, Yabiao Wang, Chengjie Wang, Ran Yi, and Lizhuang Ma.
\newblock {{RFENet}: Towards Reciprocal Feature Evolution for Glass Segmentation}.
\newblock In \emph{International Joint Conference on Artificial Intelligence}, pages 717--725, 2023{\natexlab{b}}.

\bibitem[Fu et~al.(2026)Fu, Liu, Zhou, Meng, Yang, and He]{fu2026codiff}
Xiaotong Fu, Qian Liu, Qihang Zhou, Wenchao Meng, Qinmin Yang, and Shibo He.
\newblock {{CODiff}: One-Step Diffusion Model for Camouflaged Object Detection}.
\newblock In \emph{Proceedings of the International Conference on Machine Learning}, pages 31977--31991, 2026.

\bibitem[Gao et~al.(2018)Gao, Yu, Li, Morariu, and Davis]{gao2018dynamic}
Mingfei Gao, Ruichi Yu, Ang Li, Vlad~I. Morariu, and Larry~S. Davis.
\newblock {Dynamic Zoom-In Network for Fast Object Detection in Large Images}.
\newblock In \emph{IEEE Conference on Computer Vision and Pattern Recognition}, pages 6926--6935, 2018.

\bibitem[Gao et~al.(2021)Gao, Cheng, Zhao, Zhang, Yang, and Torr]{gao2021res2net}
Shang-Hua Gao, Ming-Ming Cheng, Kai Zhao, Xin-Yu Zhang, Ming-Hsuan Yang, and Philip Torr.
\newblock {Res2Net}: A new multi-scale backbone architecture.
\newblock \emph{IEEE Transactions on Pattern Analysis and Machine Intelligence}, 43\penalty0 (2):\penalty0 652--662, 2021.

\bibitem[Han et~al.(2026{\natexlab{a}})Han, Yoon, Kwon, Kim, Choo, Lee, and Zhang]{han2024iebaf}
Dongshen Han, Heechan Yoon, Hyukmin Kwon, Hyun-Cheol Kim, Hyon-Gon Choo, Seungkyu Lee, and Chaoning Zhang.
\newblock Internal-external boundary attention fusion for glass surface segmentation.
\newblock \emph{Neural Networks}, 195:\penalty0 108232, 2026{\natexlab{a}}.

\bibitem[Han et~al.(2026{\natexlab{b}})Han, Wu, Sun, and Tang]{depthsam2026}
Jinyu Han, Changguang Wu, Fuming Sun, and Jinhui Tang.
\newblock {Beyond Appearance: Camouflaged Object Detection via Geometric Structure}.
\newblock In \emph{IEEE/CVF Conference on Computer Vision and Pattern Recognition}, pages 25830--25840, 2026{\natexlab{b}}.

\bibitem[He et~al.(2023)He, Li, Zhang, Tang, Zhang, Guo, and Li]{he2023feder}
Chunming He, Kai Li, Yachao Zhang, Longxiang Tang, Yulun Zhang, Zhenhua Guo, and Xiu Li.
\newblock {Camouflaged Object Detection With Feature Decomposition and Edge Reconstruction}.
\newblock In \emph{IEEE/CVF Conference on Computer Vision and Pattern Recognition}, pages 22046--22055, 2023.

\bibitem[He et~al.(2025)He, Zhang, Xiao, Fang, Tang, Zhang, Kong, Fan, Li, and Farsiu]{he2025run}
Chunming He, Rihan Zhang, Fengyang Xiao, Chengyu Fang, Longxiang Tang, Yulun Zhang, Linghe Kong, Deng-Ping Fan, Kai Li, and Sina Farsiu.
\newblock {{RUN}: Reversible Unfolding Network for Concealed Object Segmentation}.
\newblock In \emph{Proceedings of the International Conference on Machine Learning}, pages 22853--22864, 2025.

\bibitem[He et~al.(2026)He, Zhang, Tang, Zhang, Zhang, Xiao, Feng, and Farsiu]{he2026nun}
Chunming He, Rihan Zhang, Longxiang Tang, Dingming Zhang, Bojian Zhang, Fengyang Xiao, Jingjia Feng, and Sina Farsiu.
\newblock {Nested Unfolding Network for Real-World Concealed Object Segmentation}.
\newblock In \emph{European Conference on Computer Vision}, 2026.

\bibitem[He et~al.(2016)He, Zhang, Ren, and Sun]{he2016resnet}
Kaiming He, Xiangyu Zhang, Shaoqing Ren, and Jian Sun.
\newblock {Deep Residual Learning for Image Recognition}.
\newblock In \emph{Proceedings of the IEEE Conference on Computer Vision and Pattern Recognition (CVPR)}, pages 770--778, 2016.

\bibitem[Hu et~al.(2023)Hu, Wang, Qin, Dai, Ren, Luo, Tai, and Shao]{hu2023hitnet}
Xiaobin Hu, Shuo Wang, Xuebin Qin, Hang Dai, Wenqi Ren, Donghao Luo, Ying Tai, and Ling Shao.
\newblock {High-resolution iterative feedback network for camouflaged object detection}.
\newblock In \emph{AAAI Conf. Artif. Intell.}, pages 881--889, 2023.

\bibitem[Huang et~al.(2024)Huang, Zhou, Luo, Liu, Guo, and Yang]{huang2024ffm}
Jiaxing Huang, Yanfeng Zhou, Yaoru Luo, Guole Liu, Heng Guo, and Ge Yang.
\newblock Representing topological self-similarity using fractal feature maps for accurate segmentation of tubular structures.
\newblock In \emph{European Conference on Computer Vision}, pages 143--160, 2024.

\bibitem[Huang et~al.(2023)Huang, Dai, Xiang, Wang, Chen, Qin, and Xiong]{huang2023fspnet}
Zhou Huang, Hang Dai, Tian-Zhu Xiang, Shuo Wang, Huai-Xin Chen, Jie Qin, and Huan Xiong.
\newblock {Feature Shrinkage Pyramid for Camouflaged Object Detection With Transformers}.
\newblock In \emph{IEEE/CVF Conference on Computer Vision and Pattern Recognition}, pages 5557--5566, 2023.

\bibitem[Huynh et~al.(2021)Huynh, Tran, Luu, and Hoai]{huynh2021progressive}
Chuong Huynh, Anh~Tuan Tran, Khoa Luu, and Minh Hoai.
\newblock {Progressive Semantic Segmentation}.
\newblock In \emph{IEEE/CVF Conference on Computer Vision and Pattern Recognition}, pages 16755--16764, 2021.

\bibitem[Jaderberg et~al.(2015)Jaderberg, Simonyan, Zisserman, and Kavukcuoglu]{jaderberg2015stn}
Max Jaderberg, Karen Simonyan, Andrew Zisserman, and Koray Kavukcuoglu.
\newblock {Spatial transformer networks}.
\newblock In \emph{Adv. Neural Inform. Process. Syst.}, pages 2017--2025, 2015.

\bibitem[Jan et~al.(2025)Jan, El-Maleh, Siddiqui, Bais, and Anwar]{jan2025c3net}
Baber Jan, Aiman~H. El-Maleh, Abdul~Jabbar Siddiqui, Abdul Bais, and Saeed Anwar.
\newblock {{C3Net}: Context-Contrast Network for Camouflaged Object Detection}.
\newblock \emph{arXiv preprint arXiv:2511.12627}, 2025.

\bibitem[Jha et~al.(2020)Jha, Smedsrud, Riegler, Halvorsen, de~Lange, Johansen, and Johansen]{jha2020kvasir}
Debesh Jha, Pia~H. Smedsrud, Michael~A. Riegler, P{\aa}l Halvorsen, Thomas de Lange, Dag Johansen, and H{\aa}vard~D. Johansen.
\newblock {Kvasir-SEG}: A segmented polyp dataset.
\newblock In \emph{International Conference on Multimedia Modeling}, pages 451--462, 2020.

\bibitem[Jia et~al.(2022)Jia, Yao, Liu, Fan, Liu, and Luo]{jia2022segmar}
Qi Jia, Shuilian Yao, Yu Liu, Xin Fan, Risheng Liu, and Zhongxuan Luo.
\newblock {Segment, Magnify and Reiterate: Detecting Camouflaged Objects the Hard Way}.
\newblock In \emph{IEEE/CVF Conference on Computer Vision and Pattern Recognition}, pages 4713--4722, 2022.

\bibitem[Lamdouar et~al.(2023)Lamdouar, Xie, and Zisserman]{lamdouar2023making}
Hala Lamdouar, Weidi Xie, and Andrew Zisserman.
\newblock {The Making and Breaking of Camouflage}.
\newblock In \emph{IEEE/CVF International Conference on Computer Vision}, pages 832--842, 2023.

\bibitem[Le et~al.(2019)Le, Nguyen, Nie, Tran, and Sugimoto]{le2019camo}
Trung-Nghia Le, Tam~V Nguyen, Zhongliang Nie, Minh-Triet Tran, and Akihiro Sugimoto.
\newblock {Anabranch network for camouflaged object segmentation}.
\newblock \emph{Comput. Vis. Image Underst.}, 184:\penalty0 45--56, 2019.

\bibitem[Lin et~al.(2023)Lin, Huang, Zhou, Wang, and Zhang]{lin2023sgat}
Ji Lin, Xingru Huang, Huiyu Zhou, Yaqi Wang, and Qianni Zhang.
\newblock Stimulus-guided adaptive transformer network for retinal blood vessel segmentation in fundus images.
\newblock \emph{Medical Image Analysis}, 89:\penalty0 102929, 2023.

\bibitem[Lv et~al.(2021)Lv, Zhang, Dai, Li, Liu, Barnes, and Fan]{lv2021ranknet}
Yunqiu Lv, Jing Zhang, Yuchao Dai, Aixuan Li, Bowen Liu, Nick Barnes, and Deng-Ping Fan.
\newblock {Simultaneously Localize, Segment and Rank the Camouflaged Objects}.
\newblock In \emph{IEEE/CVF Conference on Computer Vision and Pattern Recognition}, pages 11591--11601, 2021.

\bibitem[Margolin et~al.(2014)Margolin, Zelnik-Manor, and Tal]{margolin2014wfmeasure}
Ran Margolin, Lihi Zelnik-Manor, and Ayellet Tal.
\newblock {How to Evaluate Foreground Maps?}
\newblock In \emph{Proceedings of the IEEE Conference on Computer Vision and Pattern Recognition (CVPR)}, pages 248--255, 2014.

\bibitem[Mei et~al.(2020)Mei, Yang, Wang, Liu, He, Zhang, Wei, and Lau]{mei2020gdd}
Haiyang Mei, Xin Yang, Yang Wang, Yuanyuan Liu, Shengfeng He, Qiang Zhang, Xiaopeng Wei, and Rynson W.~H. Lau.
\newblock {Don't Hit Me! Glass Detection in Real-World Scenes}.
\newblock In \emph{IEEE/CVF Conference on Computer Vision and Pattern Recognition}, pages 3687--3696, 2020.

\bibitem[Mou et~al.(2021)Mou, Zhao, Fu, Liu, Cheng, Zheng, Su, Yang, Chen, Frangi, et~al.]{mou2021cs2net}
Lei Mou, Yitian Zhao, Huazhu Fu, Yonghuai Liu, Jun Cheng, Yalin Zheng, Pan Su, Jianlong Yang, Li Chen, Alejandro~F. Frangi, et~al.
\newblock {CS2-Net}: Deep learning segmentation of curvilinear structures in medical imaging.
\newblock \emph{Medical Image Analysis}, 67:\penalty0 101874, 2021.

\bibitem[Oquab et~al.(2024)Oquab, Darcet, Moutakanni, Vo, Szafraniec, Khalidov, et~al.]{oquab2024dinov2}
Maxime Oquab, Timoth{\'e}e Darcet, Th{\'e}o Moutakanni, Huy Vo, Marc Szafraniec, Vasil Khalidov, et~al.
\newblock {DINOv2}: Learning robust visual features without supervision.
\newblock \emph{Transactions on Machine Learning Research}, 2024.

\bibitem[Pang et~al.(2022)Pang, Zhao, Xiang, Zhang, and Lu]{pang2022zoomnet}
Youwei Pang, Xiaoqi Zhao, Tian-Zhu Xiang, Lihe Zhang, and Huchuan Lu.
\newblock {Zoom in and Out: A Mixed-Scale Triplet Network for Camouflaged Object Detection}.
\newblock In \emph{IEEE/CVF Conference on Computer Vision and Pattern Recognition}, pages 2160--2170, 2022.

\bibitem[Pang et~al.(2024)Pang, Zhao, Xiang, Zhang, and Lu]{pang2024zoomnext}
Youwei Pang, Xiaoqi Zhao, Tian-Zhu Xiang, Lihe Zhang, and Huchuan Lu.
\newblock {{ZoomNeXt}: A unified collaborative pyramid network for camouflaged object detection}.
\newblock \emph{IEEE Trans. Pattern Anal. Mach. Intell.}, 46\penalty0 (12):\penalty0 9205--9220, 2024.

\bibitem[Qi et~al.(2023)Qi, He, Qi, Zhang, and Yang]{qi2023dscnet}
Yaolei Qi, Yuting He, Xiaoming Qi, Yuan Zhang, and Guanyu Yang.
\newblock Dynamic snake convolution based on topological geometric constraints for tubular structure segmentation.
\newblock In \emph{IEEE/CVF International Conference on Computer Vision}, pages 6070--6079, 2023.

\bibitem[Rahman and Marculescu(2023)]{rahman2023cascade}
Md~Mostafijur Rahman and Radu Marculescu.
\newblock Medical image segmentation via cascaded attention decoding.
\newblock In \emph{IEEE/CVF Winter Conference on Applications of Computer Vision}, pages 6222--6231, 2023.

\bibitem[Silva et~al.(2014)Silva, Histace, Romain, Dray, and Granado]{silva2014etis}
Juan Silva, Aymeric Histace, Olivier Romain, Xavier Dray, and Bertrand Granado.
\newblock Toward embedded detection of polyps in {WCE} images for early diagnosis of colorectal cancer.
\newblock \emph{International Journal of Computer Assisted Radiology and Surgery}, 9\penalty0 (2):\penalty0 283--293, 2014.

\bibitem[Skurowski et~al.(2017)Skurowski, Abdulameer, B{\l}aszczyk, Depta, Kornacki, and Kozie{\l}]{skurowski2017chameleon}
Przemys{\l}aw Skurowski, Hassan Abdulameer, Jakub B{\l}aszczyk, Tomasz Depta, Adam Kornacki, and Przemys{\l}aw Kozie{\l}.
\newblock Animal camouflage analysis: {CHAMELEON} database.
\newblock Silesian University of Technology, 2017.

\bibitem[Staal et~al.(2004)Staal, Abr{\`a}moff, Niemeijer, Viergever, and van Ginneken]{staal2004drive}
Joes Staal, Michael~D. Abr{\`a}moff, Meindert Niemeijer, Max~A. Viergever, and Bram van Ginneken.
\newblock Ridge-based vessel segmentation in color images of the retina.
\newblock \emph{IEEE Transactions on Medical Imaging}, 23\penalty0 (4):\penalty0 501--509, 2004.

\bibitem[Sun et~al.(2025)Sun, Chen, Lin, Sun, Liu, and Ji]{sun2025camodiff}
Ke Sun, Zhongxi Chen, Xianming Lin, Xiaoshuai Sun, Hong Liu, and Rongrong Ji.
\newblock {Conditional Diffusion Models for Camouflaged and Salient Object Detection}.
\newblock \emph{IEEE Transactions on Pattern Analysis and Machine Intelligence}, 47\penalty0 (4):\penalty0 2833--2848, 2025.

\bibitem[Sun et~al.(2024)Sun, Xu, Yang, Xuan, and Luo]{sun2024fsel}
Yanguang Sun, Chunyan Xu, Jian Yang, Hanyu Xuan, and Lei Luo.
\newblock {Frequency-Spatial Entanglement Learning for Camouflaged Object Detection}.
\newblock In \emph{European Conference on Computer Vision}, pages 343--360, 2024.

\bibitem[Troscianko et~al.(2017)Troscianko, Skelhorn, and Stevens]{troscianko2017quantifying}
Jolyon Troscianko, John Skelhorn, and Martin Stevens.
\newblock {Quantifying camouflage: how to predict detectability from appearance}.
\newblock \emph{BMC Evolutionary Biology}, 17\penalty0 (1):\penalty0 7, 2017.

\bibitem[Wang et~al.(2022)Wang, Xie, Li, Fan, Song, Liang, Lu, Luo, and Shao]{wang2022pvtv2}
Wenhai Wang, Enze Xie, Xiang Li, Deng-Ping Fan, Kaitao Song, Ding Liang, Tong Lu, Ping Luo, and Ling Shao.
\newblock {{PVT v2}: Improved baselines with pyramid vision transformer}.
\newblock \emph{Computational Visual Media}, 8\penalty0 (3):\penalty0 415--424, 2022.

\bibitem[Wang et~al.(2024)Wang, Sun, and Wang]{wang2024lssnet}
Wei Wang, Huiying Sun, and Xin Wang.
\newblock {{LSSNet}: A Method for Colon Polyp Segmentation Based on Local Feature Supplementation and Shallow Feature Supplementation}.
\newblock In \emph{Medical Image Computing and Computer Assisted Intervention}, pages 446--456, 2024.

\bibitem[Yan et~al.(2025)Yan, Gao, Xu, Zhu, Huang, Li, Wah, and Lau]{yan2024ghostingnet}
Tao Yan, Jiahui Gao, Ke Xu, Xiangjie Zhu, Hao Huang, Helong Li, Benjamin Wah, and Rynson W.~H. Lau.
\newblock {GhostingNet}: A novel approach for glass surface detection with ghosting cues.
\newblock \emph{IEEE Transactions on Pattern Analysis and Machine Intelligence}, 47\penalty0 (1):\penalty0 323--337, 2025.

\bibitem[Ye et~al.(2025)Ye, Chen, Zhang, Lin, and Cao]{escnet2025}
Sheng Ye, Xin Chen, Yan Zhang, Xianming Lin, and Liujuan Cao.
\newblock {ESCNet:~Edge-Semantic Collaborative Network for Camouflaged Object Detection}.
\newblock In \emph{IEEE/CVF International Conference on Computer Vision}, pages 20053--20063, 2025.

\bibitem[Yin et~al.(2024)Yin, Zhang, Fan, Jiao, Cheng, Van~Gool, and Hou]{yin2024camoformer}
Bowen Yin, Xuying Zhang, Deng-Ping Fan, Shaohui Jiao, Ming-Ming Cheng, Luc Van~Gool, and Qibin Hou.
\newblock {{CamoFormer}: Masked separable attention for camouflaged object detection}.
\newblock \emph{IEEE Trans. Pattern Anal. Mach. Intell.}, 46\penalty0 (12):\penalty0 10362--10374, 2024.

\bibitem[Yin et~al.(2025)Yin, Zhang, Liu, Cheng, Liu, and Hou]{yin2025adaptcod}
Bowen Yin, Xuying Zhang, Li Liu, Ming-Ming Cheng, Yongxiang Liu, and Qibin Hou.
\newblock {Camouflaged Object Detection with Adaptive Partition and Background Retrieval}.
\newblock \emph{International Journal of Computer Vision}, 133:\penalty0 4877--4893, 2025.

\end{thebibliography}


\begin{thebibliography}{4}
\providecommand{\natexlab}[1]{#1}
\providecommand{\url}[1]{\texttt{#1}}
\expandafter\ifx\csname urlstyle\endcsname\relax
  \providecommand{\doi}[1]{doi: #1}\else
  \providecommand{\doi}{doi: \begingroup \urlstyle{rm}\Url}\fi

\bibitem[Fan et~al.(2022)Fan, Ji, Cheng, and Shao]{fan2022concealed}
Deng-Ping Fan, Ge-Peng Ji, Ming-Ming Cheng, and Ling Shao.
\newblock {Concealed object detection}.
\newblock \emph{IEEE Trans. Pattern Anal. Mach. Intell.}, 44\penalty0 (10):\penalty0 6024--6042, 2022.

\bibitem[He et~al.(2023)He, Li, Zhang, Tang, Zhang, Guo, and Li]{he2023feder}
Chunming He, Kai Li, Yachao Zhang, Longxiang Tang, Yulun Zhang, Zhenhua Guo, and Xiu Li.
\newblock {Camouflaged Object Detection With Feature Decomposition and Edge Reconstruction}.
\newblock In \emph{IEEE/CVF Conference on Computer Vision and Pattern Recognition}, pages 22046--22055, 2023.

\bibitem[Margolin et~al.(2014)Margolin, Zelnik-Manor, and Tal]{margolin2014wfmeasure}
Ran Margolin, Lihi Zelnik-Manor, and Ayellet Tal.
\newblock {How to Evaluate Foreground Maps?}
\newblock In \emph{Proceedings of the IEEE Conference on Computer Vision and Pattern Recognition (CVPR)}, pages 248--255, 2014.

\bibitem[Sun et~al.(2024)Sun, Xu, Yang, Xuan, and Luo]{sun2024fsel}
Yanguang Sun, Chunyan Xu, Jian Yang, Hanyu Xuan, and Lei Luo.
\newblock {Frequency-Spatial Entanglement Learning for Camouflaged Object Detection}.
\newblock In \emph{European Conference on Computer Vision}, pages 343--360, 2024.

\end{thebibliography}
\end{document}


\AddToShipoutPicture{%
  \AtTextUpperLeft{%
    \put(0,\LenToUnit{36pt}){%
      \normalfont\rmfamily\small\color{black}Preprint. Under review.}%
    \put(0,\LenToUnit{32.6pt}){\color{black}\rule{\textwidth}{.4pt}}%
  }%
}
\raggedbottom
\setlength{\abovedisplayskip}{5pt plus 1pt minus 1pt}
\setlength{\belowdisplayskip}{5pt plus 1pt minus 1pt}
\setlength{\abovedisplayshortskip}{3pt}
\setlength{\belowdisplayshortskip}{3pt}
\setlength{\intextsep}{7pt plus 1pt minus 1pt}
\raggedend
\makeatletter
\renewcommand\paragraph{\@startsection{paragraph}{4}{\z@}{5pt plus 1pt minus 1pt}{-1em}{\normalfont\normalsize\bfseries}}
\makeatother
\maketitlesupplementary
\appendix
\section{Inference Procedure}\label{sup:inference}
For image $I$, supports are predicted foreground components; $Z$ stores segmentation logits and $F$ stores feature memory. The shared recognizer $\mathcal R_\theta$ uses global foreground/background references $\mathcal G_0$. Localization $O$, view selection $V$, and benefit predictor $G$ propose and score candidate windows $\mathcal C$ from concealment spectra.

For crop $I|_{V_j}$, superscript $\mathrm{loc}$ denotes local outputs. Preference $Q$ sets spatial acceptance $\lambda$; $\Pi_V$ projects to the receiving grid and $\rho_V$ gives pixel coverage. Here $\sigma$ is sigmoid and $\odot$ is elementwise multiplication.

\begingroup
\setlength{\intextsep}{5.8pt}
\begin{algorithm}[H]
\caption{Concealment-guided inference}\label{sup:algorithm}
\begin{algorithmic}[1]
\Require Image $I$, trained model, cap $T_{\max}$, thresholds $\{\zeta_t\}$
\Ensure Foreground probabilities $\sigma(Z)$
\State Initialize $Z,F,\mathcal G_0$ and object supports from $I$
\For{$t=1,\ldots,T_{\max}$}
\State Measure concealment spectra on current supports
\State Propose and score $\mathcal C$ with $O,V,G$
\StopIf{$\mathcal C=\varnothing$}
\State $j\gets\argmax_{j\in\mathcal C}G(j)$
\StopIf{$G(j)\leq\zeta_t$}
\State $(Z^{\rm loc},F^{\rm loc})\gets\mathcal R_\theta(I|_{V_j},\mathcal G_0)$
\State Compute object-conditioned preference $Q$
\State $\lambda\gets\rho_{V_j}\odot\sigma(\Pi_{V_j}Q)$
\State $Z\gets Z+\lambda\odot(\Pi_{V_j}Z^{\rm loc}-Z)$
\State Update $F$, supports, and observation histories
\EndFor
\State \Return $\sigma(Z)$
\end{algorithmic}
\end{algorithm}
\endgroup

We use $T_{\max}=3$ and a per-track cap of three. Candidates require valid windows and finite gains. Thresholds $\zeta_t$ are $-.005$, $-.0025$, and $-.00225$ at successive steps.

\paragraph{Window geometry.}
Logits $a,b$ set normalized length $\ell$ and origin $l$ along each axis of an $H\times W$ image:
\begin{equation}
\ell=\ell_{\min}+(1-\ell_{\min})\sigma(a),\qquad l=(1-\ell)\sigma(b),
\end{equation}
where $\ell_{\min}=1/W$ horizontally and $1/H$ vertically.

\paragraph{Support updates.}
Otsu thresholding on a $352\times352$ grid defines eight-connected supports. Tracks match when intersection divided by the smaller area is at least $.2$. Splits share histories; merges combine histories and retain the maximum count; unmatched components start new tracks. Fusion updates memory and remeasures affected supports and backgrounds while retaining out-of-view histories.

\begin{table*}[!t]
\centering
\caption{\textbf{Transfer of \method to existing COD detectors.}}
\label{sup:detector_transfer}
\compacttable\setlength{\tabcolsep}{1.5pt}
\columntable{17}{%
\begin{tabular}{l|cccc|cccc|cccc|cccc}\toprule
& \multicolumn{4}{c|}{CHAMELEON} & \multicolumn{4}{c|}{CAMO} & \multicolumn{4}{c|}{COD10K} & \multicolumn{4}{c}{NC4K}\\
\rowcolor{tablehead}Method & $S_\alpha\uparrow$ & $F_\beta\uparrow$ & $E_\phi\uparrow$ & $M\downarrow$ & $S_\alpha\uparrow$ & $F_\beta\uparrow$ & $E_\phi\uparrow$ & $M\downarrow$ & $S_\alpha\uparrow$ & $F_\beta\uparrow$ & $E_\phi\uparrow$ & $M\downarrow$ & $S_\alpha\uparrow$ & $F_\beta\uparrow$ & $E_\phi\uparrow$ & $M\downarrow$\\\midrule
FEDER~\cite{he2023feder} & .8949 & .8480 & \textbf{.9506} & .0282 & .7824 & .7449 & .8413 & .0799 & .8014 & .6963 & .8738 & .0368 & .8341 & .7947 & .8921 & .0490\\
\rowcolor{tableours}\quad +\method & \textbf{.8970} & \textbf{.8505} & .9501 & \textbf{.0278} & \textbf{.7855} & \textbf{.7471} & \textbf{.8419} & \textbf{.0792} & \textbf{.8047} & \textbf{.7010} & \textbf{.8746} & \textbf{.0361} & \textbf{.8360} & \textbf{.7958} & \textbf{.8924} & \textbf{.0485}\\
\midrule
FSEL~\cite{sun2024fsel} & .8947 & .8450 & .9463 & .0285 & .7831 & .7485 & \textbf{.8422} & .0807 & .8005 & .6930 & \textbf{.8714} & .0374 & .8342 & .7954 & \textbf{.8905} & .0494\\
\rowcolor{tableours}\quad +\method & \textbf{.8978} & \textbf{.8488} & \textbf{.9489} & \textbf{.0277} & \textbf{.7846} & \textbf{.7498} & .8395 & \textbf{.0800} & \textbf{.8036} & \textbf{.7041} & \textbf{.8714} & \textbf{.0362} & \textbf{.8347} & \textbf{.7996} & .8886 & \textbf{.0489}\\
\midrule
SINet-V2~\cite{fan2022concealed} & .8900 & .8312 & .9368 & .0317 & .8124 & .7754 & .8667 & .0713 & .8213 & .7039 & .8864 & .0340 & .8476 & .7998 & .9006 & .0460\\
\rowcolor{tableours}\quad +\method & \textbf{.8966} & \textbf{.8544} & \textbf{.9445} & \textbf{.0288} & \textbf{.8145} & \textbf{.7856} & \textbf{.8698} & \textbf{.0694} & \textbf{.8293} & \textbf{.7363} & \textbf{.8942} & \textbf{.0312} & \textbf{.8497} & \textbf{.8139} & \textbf{.9032} & \textbf{.0445}\\
\bottomrule\end{tabular}}
\end{table*}

\newpage
\section{Concealment Measurement}\label{sup:measurement}
\begin{figure}[!ht]
\centering
\includegraphics[width=\columnwidth]{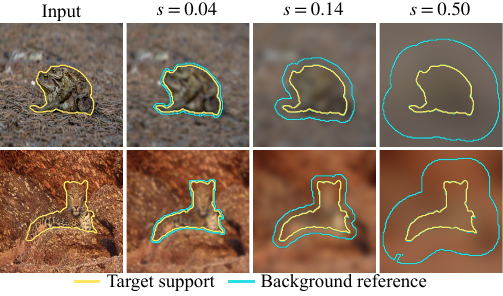}
\caption{\textbf{Camouflage cues across scales.} Yellow outlines the target; cyan marks the outer boundary of its background reference.}
\label{sup:measurement_scales}
\end{figure}
In Fig.~\ref{sup:measurement_scales}, fine patterns blend with the surroundings at small $s$. Increasing $s$ suppresses these textures and exposes body--background tonal differences, but further smoothing erases contour detail. These cues favor different scales and jointly guide revealing local observations.

\paragraph{Measurement settings.}
\mbox{$K=9$} logarithmic scales cover $.04$--$.50$ of equivalent target diameter $D$. We use a \mbox{$128\times128$} grid and correction bound \mbox{$\alpha_c=.3$}. Descriptors comprise smoothed intensity, red--green and green--blue differences, gradient magnitude, local standard deviation, and deviation from the local mean. Bandwidth is $h_k=s_kD$; the ring spans $\max(1,h_k)$ pixels outside foreground. Covariances use diagonal shrinkage and a $.03^2I$ regularizer.

Gaussian-weighted boundary color contrast excludes crop edges and other foreground. We use $\tau_{\rm col}=\sqrt3$, $\epsilon_B=.05$, $\kappa=-4$, and $\epsilon_\Delta=.05$. Missing evidence takes $.5$ and zero reliability, including empty foreground or background. Gain and preference use $\sum_k\max(-d_k,0)$ for rising concealment toward coarser scales.

\paragraph{Measurement supervision.}
The probe takes annotated-object \mbox{$128\times128$} patches, masks, foreground unions, and geometry. Formula targets preserve aspect ratio (maximum side 256). Reliability-weighted absolute error supervises curves; squared error and smooth-$L_1$ supervise reliability and curve-error prediction, each weighted $.2$. Predicted error attenuates confidence. Correction aligns measurements on predicted training supports with aspect-preserving targets using absolute losses on values, mean-centered curves, and adjacent-scale differences (weights $1$, $.5$, $1$).

\Needspace{8\baselineskip}
\section{Transfer to Existing Detectors}
We add \method to FEDER and FSEL (\mbox{ResNet-50}) and SINet-V2 (\mbox{Res2Net-50}). The detectors train for 180 epochs using their official implementations and losses. Their weights and BatchNorm remain fixed while the added observation modules train for 24 epochs.

Across all 12 detector--dataset pairs, \method improves $F_\beta$, $S_\alpha$, and MAE (Table~\ref{sup:detector_transfer}). SINet-V2 gains 2.32, 1.03, 3.24, and 1.41 percentage points in $F_\beta$ on CHAMELEON, CAMO, COD10K, and NC4K, respectively, supporting transfer across the tested detectors.

\section{Training Details}\label{sup:training}
\paragraph{Prediction supervision.}
Recognition uses boundary-weighted BCE and IoU together with Dice. Observation learning weights integrated error by $1$ and local error by $.5$, using the full-image error at native resolution:
\begin{equation}
\mathcal E(z,y)=1-F^w_1(p,y)+.1\operatorname{BCE}(p,y),\quad p=\sigma(z),
\label{sup:eq:native_error}
\end{equation}
where $z$ denotes logits, $y$ the label, and $F^w_1$ weighted $F$-measure~\cite{margolin2014wfmeasure}. For local error, projected logits replace the current logits inside the crop, retaining predictions outside. Structure losses match foreground/background statistics and spatial layout: global and integrated predictions each use weight $.25$, and local predictions use $.125$.

Localization learns target extent using generalized IoU and smooth-$L_1$. View selection uses balanced BCE and Dice on the integrated prediction, with complete matched targets and in-crop background as supervision.

Preference compares the current and local predictions against the label, rewarding foreground recovery and penalizing distractors. Its balanced BCE uses prediction disagreement and crop coverage as pixel weights, with overall coefficient $.05$. Gain predicts the reduction in $\mathcal E$ after observation; smooth-$L_1$ has weight $.1$, with predicted gains and observed reductions both divided by $.05$.

\paragraph{Optimization.}
The recognizer uses Adam for \mbox{ResNet-50} and AdamW otherwise, with three warmup epochs followed by cosine decay. Concealment-conditioned modules use AdamW with weight decay $10^{-4}$ during adaptation. Measurement correction uses AdamW with an initial learning rate of $5\times10^{-4}$ and a weight decay of $.01$.

\newpage
\section{Failure Cases}
Camouflage can obscure extent or identity, leaving supports incomplete or misplaced and biasing later measurements and view selection. Local foreground/background similarity can also leave recognition ambiguous, allowing residual boundary errors to persist through selective fusion.

In Fig.~\ref{sup:failure_cases} (row~2), a striped distractor's color contrast dominates the prediction while the target remains missed. The ghost pipefish shares texture with the adjacent rock (row~4), leaving connected foreground after fusion.

\begingroup
\setlength{\intextsep}{1pt}
\begin{figure}[!ht]
\centering
\includegraphics[width=\columnwidth]{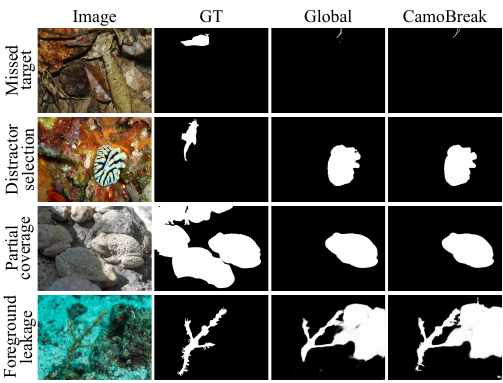}
\caption{\textbf{Residual errors after local observation on COD10K.}}
\label{sup:failure_cases}
\end{figure}
\FloatBarrier
\endgroup
\vspace{-11.27pt}
\raggedend

{\small\bibliographystyle{ieeenat_fullname}\bibliography{main}}